\documentclass[final]{cvpr}

\usepackage{times}
\usepackage{epsfig}
\usepackage{graphicx,subfigure}
\usepackage{amsmath}
\usepackage{amssymb}
\usepackage{url}
\usepackage{mathrsfs}
\usepackage{algorithmicx,algorithm}
\usepackage{verbatim}
\usepackage{amsfonts,amssymb}
\usepackage{bm}
\usepackage{mathtools}
\usepackage{makecell, multirow}
\usepackage{color}
\usepackage{pdfpages}

\usepackage{caption}
\usepackage{booktabs}

\newtheorem{theorem}{Theorem}

\usepackage[pagebackref=true,breaklinks=true,colorlinks,bookmarks=false]{hyperref}

\def\cvprPaperID{7739} % *** Enter the CVPR Paper ID here
\def\confYear{CVPR 2021}

\begin{document}

%%%%%%%%% TITLE
\title{Relaxed Conditional Image Transfer for Semi-supervised Domain Adaptation}

% \author[1]{Qijun Luo}
% \author[2]{Zhili Liu}

% \affil[1]{Chinese University of Hong Kong, Shen Zhen}
% \affil[2]{Huawei Noah's Ark Lab}

\author{
$\text{Qijun Luo}^{1\footnotemark[1]},
\text{Zhili Liu}^{2,3 \footnotemark[1]}, 
\text{Lanqing Hong}^{3}, 
\text{Chongxuan Li}^{4\dagger}, 
\text{Kuo Yang}^{3},$ \\
$\text{Liyuan Wang}^{4}, 
\text{Fengwei Zhou}^{3}, 
\text{Guilin Li}^{3}, 
\text{Zhenguo Li}^{3}, 
\text{Jun Zhu}^{4\dagger}$\\
$\prescript{1}{}{\text{Chinese University of Hong Kong, Shen Zhen, China}}$\\
$\prescript{2}{}{\text{The Hong Kong University of Science and Technology, China}}$\\
$\prescript{3}{}{\text{Huawei Noah's Ark Lab, China}}$\\
$\prescript{4}{}{\text{Dept. of Comp. Sci. $\&$ Tech., Institute for AI, THBI Lab, BNRist Center,}}$\\
$\text{State Key Lab for Intell. Tech. $\&$ Sys., Tsinghua University, Beijing, China}$
}
%\tt\small $\dagger$ chongxuanli1991@gmail.com
% {\tt\small qijunluo@link.cuhk.edu.cn, \\
% \tt\small zliudj@connect.ust.hk, \\
% \tt\small\{chongxuanli1991, hiliguilin\}@gmail.com, \\
% \tt\small\{honglanqing, yang.kuo, zhoufengwei, li.zhenguo\}@huawei.com, \\ \tt\small wly19@mails.tsinghua.edu.cn,
% \tt\small
% dcszj@mail.tsinghua.edu.cn}

\maketitle

\renewcommand{\thefootnote}{\fnsymbol{footnote}}
\footnotetext[1]{These authors contribute equally to this work. The work is done when Qijun Luo and Zhili Liu are interns in Huawei Noah's Ark Lab. $\dagger$ Corresponding authors: C. Li, email: chongxuanli1991@gmail.com, and J. Zhu, email: dcszj@mail.tsinghua.edu.cn.}

%%%%%%%%% ABSTRACT
% not emphasize label domination
% not emphasize theoretical proof
% mention cGAN -> but identify label domination

% prelimi: cGAN + cGAN loss
% Method
    % label domination
        % GD
        % path angle: comp with preliminary, indicate better practical convergence.
    % other factors
        % C + marg + cyc ... 
        % equilibrium
    
% problem, solu, exp, visual
% other factor(marginal + joint C)

% abstract
    % SSDA setting
    % label inconsistency -> cGAN -> label domination -> rcGAN -> equilibrium analysis & convergence -> good image transformation & classification -> additional enhancement for SSDA -> experiment summary (digit, domainnet, office)

\begin{abstract}
% Among domain adaptation methods, image transfer is a popular technique. However, it often suffers from the label-inconsistency problem, where the generated image's label is different from the original image, resulting in a poor adaptation. In SSDA, a straightforward way to address this is to adopt a conditional GAN framework to capture the class-conditional distribution of the generated image. 
Semi-supervised domain adaptation (SSDA), which aims to learn models in a partially labeled target domain with the assistance of the fully labeled source domain,
attracts increasing attention in recent years.
To explicitly leverage the labeled data in both domains, we naturally introduce a conditional GAN framework to transfer images without changing the semantics in SSDA.
However, we identify a label-domination problem in such an approach. In fact, the generator tends to overlook the input source image and only memorizes prototypes of each class, which results in unsatisfactory adaptation performance. To this end, we propose a simple yet effective Relaxed conditional GAN (Relaxed cGAN) framework. Specifically, we feed the image without its label to our generator. In this way, the generator has to infer the semantic information of input data. We formally prove that its equilibrium is desirable and empirically validate its practical convergence and effectiveness in image transfer. Additionally, we propose several techniques to make use of unlabeled data in the target domain, enhancing the model in SSDA settings. We validate our method on the well-adopted datasets: Digits, DomainNet, and Office-Home. We achieve state-of-the-art performance on DomainNet, Office-Home and most digit benchmarks in low-resource and high-resource settings.

% Under the SSDA settings, it is important to provide appropriate image transformation from the source domain to the target one.
% However, such image transfer often encounters two problems, namely, the \textit{label-inconsistency} problem and the \textit{label-domination} problem, where information of the source image is ignored during adaptation.
% In this paper, we propose a GAN-based approach, called \textbf{Relaxed} \textbf{c}onditional \textbf{GAN} (Relaxed cGAN), to address these two problems, improve the image transfer quality, and eventually benefit the classification accuracy.
% In particular, we introduce a conditional discriminator which discriminates label-data pairs in the target domain to solve the label-inconsistency problem. 
% More importantly, we further adopt an unconditional generator that takes the source image as the only input.
% In this way, the generator has to infer the semantic information of input data. 
% We empirically demonstrate that such design significantly alleviate the label-domination problem.
% Additionally, we consider a jointly trained classifier to make use of both labeled data and unlabeled data in the target domain. 
% We validate our method in the widely adopted digit and DomainNet adaptation benchmarks, achieving state-of-the-art performance in digit benchmarks and competitive results in DomainNet under both low-resource and high-resource SSDA settings.
	
\end{abstract}

\section{Introduction}
\label{sec:intro}

% [SSDA is important]
While deep neural networks (DNNs) have made remarkable progress on various visual tasks~\cite{he2016deep, krizhevsky2012imagenet, szegedy2015going}, they usually rely on a large amount of labeled data in model training.
However, it is often costly to collect extensive training data with labels for a new coming task.
To alleviate the burden of data collection, 
domain adaptation (DA) methods aim to learn a model on a target domain with the assistance of a source domain in which sufficient annotated data are available~\cite{pan2009survey}.
Existing DA methods~\cite{ganin2016domain, tzeng2017adversarial, tang2020unsupervised, long2017deep, bhushan2018deepjdot, kang2019contrastive} mainly focus on unsupervised settings, where labels are unavailable in the target domain.
Some recent works, however, have shown that a few labeled data in the target domain could significantly boost the model performance~\cite{saito2018maximum,motiian2017few,hosseini2018augmented}.
Considering that it is usually acceptable to collect a few labeled data of a target task for performance improvement, 
semi-supervised domain adaptation (SSDA) attracts more and more attention in recent years, which investigates to learn models in a partially labeled target domain with fully labeled data from a related source domain~\cite{motiian2017few,hosseini2018augmented,perez2019matching}.

% [Categorize the existing method into two main categories: feature alignment and transformation]

Current SSDA methods mainly fall into two categories, namely, the feature-level adaptation and the pixel-level adaptation. 
For feature-level adaptation, most methods either learn an invariant representation across domains by minimizing certain discrepancy~\cite{zou2019consensus} or extract target discriminative features with source domain as regularization~\cite{saito2019semi}.
For pixel-level adaptation, the main idea is to transfer source domain images to the target domain's style, serving as data augmentation~\cite{hosseini2018augmented}. 
In this paper, we mainly focus on pixel-level SSDA methods, as it enjoys the advantages of more stable model training and higher interpretability as the transferred images can be visualized and re-used~\cite{bousmalis2017unsupervised}.

% Compared with feature-level approaches, pixel-level adaptation enjoys the advantages of more stable model training and higher interpretability as the transferred images can be visualized and re-used~\cite{bousmalis2017unsupervised}. Based on these considerations, we mainly focus on pixel-level SSDA methods. 

Among pixel-level adaptation, the cycle-consistency loss is a popular technique for image transformation~\cite{zhu2017unpaired, hoffman2017cycada, hosseini2018augmented, murez2018image, luo2019taking, liu2017unsupervised, huang2018auggan}.
Such an approach, however, often suffers from a major obstacle named label-inconsistency, in which the generated image's label is different from the input image~\cite{hosseini2018augmented, murez2018image, luo2019taking}. To address this issue in SSDA, a straightforward and common way is to adopt a conditional GAN to align the class-conditional distribution of generated image-label pairs. Nevertheless, we empirically show that such design meets a \emph{label-domination} problem, in which the generator tends to neglect the source image's information and only generates images according to the input label (see Fig.~\ref{fig:label-domination} for empirical evidence). Such an issue would result in unsatisfactory adaptation due to the ineffective usage of the source domain's information. 

To alleviate this issue, we propose \textbf{Relaxed} \textbf{c}onditional \textbf{GAN} (Relaxed cGAN), where a source image is fed into the generator for adaptation \textit{without} the label. 
The resulting generated image together with the true label of the input image needs to fool the conditional discriminator.
In this way, the generator is enforced to learn useful features from the raw images and generate meaningful samples in the target domain. We show that the proposed method has provable theoretical equilibrium and practical convergence properties. Moreover, we empirically demonstrates satisfactory image quality in transformation (see Fig.~\ref{fig:rcgan-generation}). 

We compare Relaxed cGAN to several strong baselines, including both feature-based ones~\cite{perez2019matching, zou2019consensus, motiian2017few, motiian2017unified, saito2019semi} and pixel-based ones~\cite{hosseini2018augmented, zhu2017unpaired}, on the commonly used Digit and more challenging DomainNet and Office-Home datasets. 
We obtain state-of-the-art results on most Digit datasets and DomainNet. Digits are conducted in both low-resource and high-resource settings. Competitive results are also obtained on the Office-Home.

To summarize, our main contributions are:
\begin{itemize}
	\item We identify the label-domination problem on a natural and widespread conditional GAN framework for SSDA.
	\item We propose Relaxed cGAN, with carefully designed modules and loss functions to address the label-domination problem and validate our intuition in both theory and practice. Besides, we propose a marginal loss to enhance Relaxed cGAN without affecting the theoretical results.
	\item We obtain promising results in various SSDA settings. As a pixel-level method, we obtain state-of-the-art performance on widely adopted Digit datasets and DomainNet, and competitive results on Office-Home.
\end{itemize}

\section{Related Work}
\label{gen_inst}

\noindent \textbf{Unsupervised Domain Adaptation.}
Unsupervised domain adaptation (UDA) aims to transfer knowledge from a source domain to a totally unlabeled target domain.
Existing UDA methods mainly fall into two categories. One is the feature-level UDA algorithms, which propose to learn a domain-invariant representation by minimizing certain discrepancy measurements. A lot of measurements are defined by maximum mean discrepancy~\cite{gretton2007kernel, rozantsev2018beyond, long2017deep}, Wasserstein metric~\cite{arjovsky2017wasserstein}, graph matching~\cite{das2018graph}, and adversarial methods~\cite{ganin2016domain, tzeng2017adversarial}. 
Another category adapts to domains in pixel-level, transferring samples from the source domain to the target domain's style~\cite{hoffman2017cycada, bousmalis2017unsupervised, murez2018image, deng2018image}. Such methods are mostly based on the cycle-consistency loss ~\cite{zhu2017unpaired}.
Although the UDA settings are widely considered, it is usually affordable to collect a few labeled data in the target domain to boost the model performance. 
Therefore, SSDA may become a more practical setting.

\noindent \textbf{Semi-supervised Domain Adaptation.}
As mentioned above, the target domain may have partially labeled data in many applications, which is referred to as the SSDA setting. There are a few methods that explore the setting.
% ~\cite{saito2018maximum,hosseini2018augmented,perez2019matching,saito2019semi} 
% may not provide performance improvement and sometimes even make it worse~\cite{saito2019semi}. 
In feature-level, ~\cite{saito2019semi} utilizes unlabeled target domain's data by minimizing their distance with class prototypes through a minimax training. Other approaches include mapping samples into shared embedding~\cite{perez2019matching} and greedily minimizing information entropy~\cite{zou2019consensus}.
On the pixel space, \cite{hosseini2018augmented} imposes a classifier to serve as a complementary discriminator. However, it doesn't utilize the target domain unlabeled data well.
% Previous methods~\cite{saito2018maximum,hosseini2018augmented,perez2019matching,saito2019semi} often extend UDA methods by utilizing the limited number of target labels.
In comparison, our Relaxed cGAN proposes to achieve better image transformation with the help of both the labeled and unlabeled data in the target domain, and eventually benefits the clarification accuracy. 

\section{A Preliminary Framework}
\label{sec:preliminary}
% \zhili{better not use preliminary in section name}

We first formulate semi-supervised domain adaptation (SSDA) and introduce a preliminary framework based on conditional GAN~\cite{mirza2014conditional} in Sec.~\ref{subsec:setting-framework}. Then, we identify a \emph{label-domination} problem of the framework in Sec.~\ref{subsec:label-dom}.  

\subsection{Settings and Framework}
\label{subsec:setting-framework}
% outline SSDA problem
In the SSDA settings, the source domain contains a large set of labeled data  $\mathcal{D}_{s}=\left\{\left(x_{i}^{s}, {y_{i}}^{s}\right)\right\}_{i=1}^{m_{s}}$.  
In the target domain, we have a small set of labeled data $\mathcal{D}_{t}=\left\{\left(x_{i}^{t}, {y_{i}}^{t}\right)\right\}_{i=1}^{m_{t}}$ and a large set of unlabeled data $\mathcal{D}_{u} = \left\{\left(x_{i}^{u} \right)\right\}_{i=1}^{m_{u}}$. 
Let $p_s(x, y)$, $p_t(x, y)$ and $p_u(x)$ denote the empirical distributions over $\mathcal{D}_{s}, \mathcal{D}_{t}, \text{and } \mathcal{D}_{u}$, respectively. 
$p_s(x)$ and $p_t(x)$ denote the marginal distributions of $x$ with respect to $p_s(x, y)$ and $p_t(x, y)$, respectively.

% cGAN + cycleGAN: a straight-forward method for ensuring label consistency.

\begin{figure}[t]
    \centering
    % \subfigure[CycleGAN]{\includegraphics[height = 2.5cm]{figures/Fig2b_n.png}}
    % \subfigure[Relaxed cGAN]{\includegraphics[height = 2.5cm]{figures/Fig2a_n.png}}
    \includegraphics[height = 2.5cm]{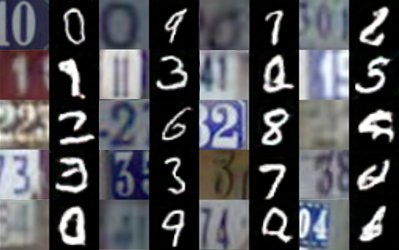}
    \caption{Label-inconsistency problem. Each odd column shows the input source images, and its next column are corresponding generated images in the target domain. CycleGAN suffers from the severe label-inconsistency problem in most cases.}
    \label{fig:label-inconsistency}
\end{figure}

To adapt to different domains, cycle consistency constraint is a popular technique. It transfers the image in the source domain to match the target domain's style by ensuring that the generated image can be reconstructed back to the original one. The generated image accompanied by the original label can serve as the augmentation of annotated target images. 
However, when the domain shift is large, such an approach may encounter the label-inconsistency problem~\cite{hoffman2017cycada} where the label of the generated image is different from the input one.
For instance, when transferring the image from SVHN~\cite{netzer2011reading} to MNIST~\cite{lecun1998gradient}, the original image of label ``5" may be transferred to a target image of label ``3", as shown in Fig~\ref{fig:label-inconsistency}(b). This could happen because there does not exist any inductive bias that ensures the label consistency should be maintained during the transformation. 

A natural way to address this issue is to impose conditional GAN~\cite{mirza2014conditional} to transfer images so that the class-conditional distribution between generated images and target domain images can be aligned. Specifically, we design a preliminary framework which adopts a conditional generator and a conditional discriminator with the following loss:
\begin{equation*}
\begin{aligned}
\mathcal{L}_{GAN} &= \mathbb{E}_{(x,y) \sim {p}_{t}(x,y)}[\log (D_{T}(x,y))]  \\
&+ \mathbb{E}_{(x,y) \sim {p}_{s}(x,y)}[\log(1- D_{T}(G_{S\rightarrow T}(x, y), y))], \\
% &+\mathbb{E}_{(x,y) \sim {p}_{s}(x,y)}[\log (D_{S}(x,y))]  \\
% &- \mathbb{E}_{(x,y) \sim {p}_{t}(x,y)}[\log(1- D_{S}(G_{T\rightarrow S}(x, y), y))], \\
\end{aligned}
\end{equation*}
where $G_{S\rightarrow T}$ is the conditional generator from source to target. $D_T$ is the conditional discriminator of the target domain, distinguishing whether the generated image accompanied by the input label is close to the desired joint distribution. For the ease of later discussion, we omit the loss of the opposite direction from target to source, which also contains a conditional generator and a conditional discriminator. To maintain the correlation between the source image and the generated image, we also apply conventional cycle consistency loss, which is formulated as 
\begin{equation*}
\begin{aligned}
\mathcal{L}_{cycle} &= \mathbb{E}_{(x,y) \sim {p}_{s}(x,y)}[|G_{T\rightarrow S}(G_{S\rightarrow T}(x,y),y)-x|] \\
&+ \mathbb{E}_{(x,y) \sim {p}_{t}(x,y)}[|G_{S\rightarrow T}(G_{T\rightarrow S}(x,y),y)-x)|],
\end{aligned}
\end{equation*}
where $|\cdot|$ is the L1 distance on reconstruction error.
To train a classifier, we apply standard cross-entropy loss on both target annotated images and generated images.
The corresponding loss is 
\begin{equation}
\begin{aligned}
\mathcal{L}_{C} &= \mathbb{E}_{(x,y) \sim p_s(x,y)}[-\log(C(y|G_{S \rightarrow T}(x)))] \\
&+ \mathbb{E}_{(x,y) \sim p_t(x,y)}[-\log(C(y|x))],
% &+ \frac{1}{K}\mathbb{E}_{x \sim p_u(x)}[\sum_{i=1}^{K}-C(i|x)\log C(i|x)]. \\
\label{eq:pre-l_c}
\end{aligned}
\end{equation}
where $C(y|x)$ is the $y_{th}$ class's probability predicted by the classifier for image $x$.

The overall objective function of our preliminary framework would be
\begin{equation*}
\mathcal{L} = \lambda_{GAN}\mathcal{L}_{GAN} + \lambda_{cycle}\mathcal{L}_{cycle} + \lambda_{C}\mathcal{L}_{C},
\end{equation*}
where $\lambda_{GAN}, \lambda_{cycle}, \lambda_{C}$ are the corresponding weight for these losses.
% where $G_{T\rightarrow S}$ and $G_{S\rightarrow T}$ are the generator from target to source domain, and source to target domain, respectively.  $D_S$ and $D_T$ are the discriminator of source and target domain, respectively. $C$ is the classifier of target domain and $C(y|x)$ denotes the corresponding conditional probability. $\mathcal{L}_{GAN}$ is the standard loss for conditional GAN~\cite{}, except that we change the latent vector into a image. $\mathcal{L}_{cycle}$ is the cycle consistency loss, and $\mathcal{L}_C$ is the loss for classifier, including the commonly used cross-entropy loss and the 
% %explain G, D

\subsection{The Label-domination Problem}
\label{subsec:label-dom}

\begin{figure}[tbp]  
	\centering
	\includegraphics[height = 3cm]{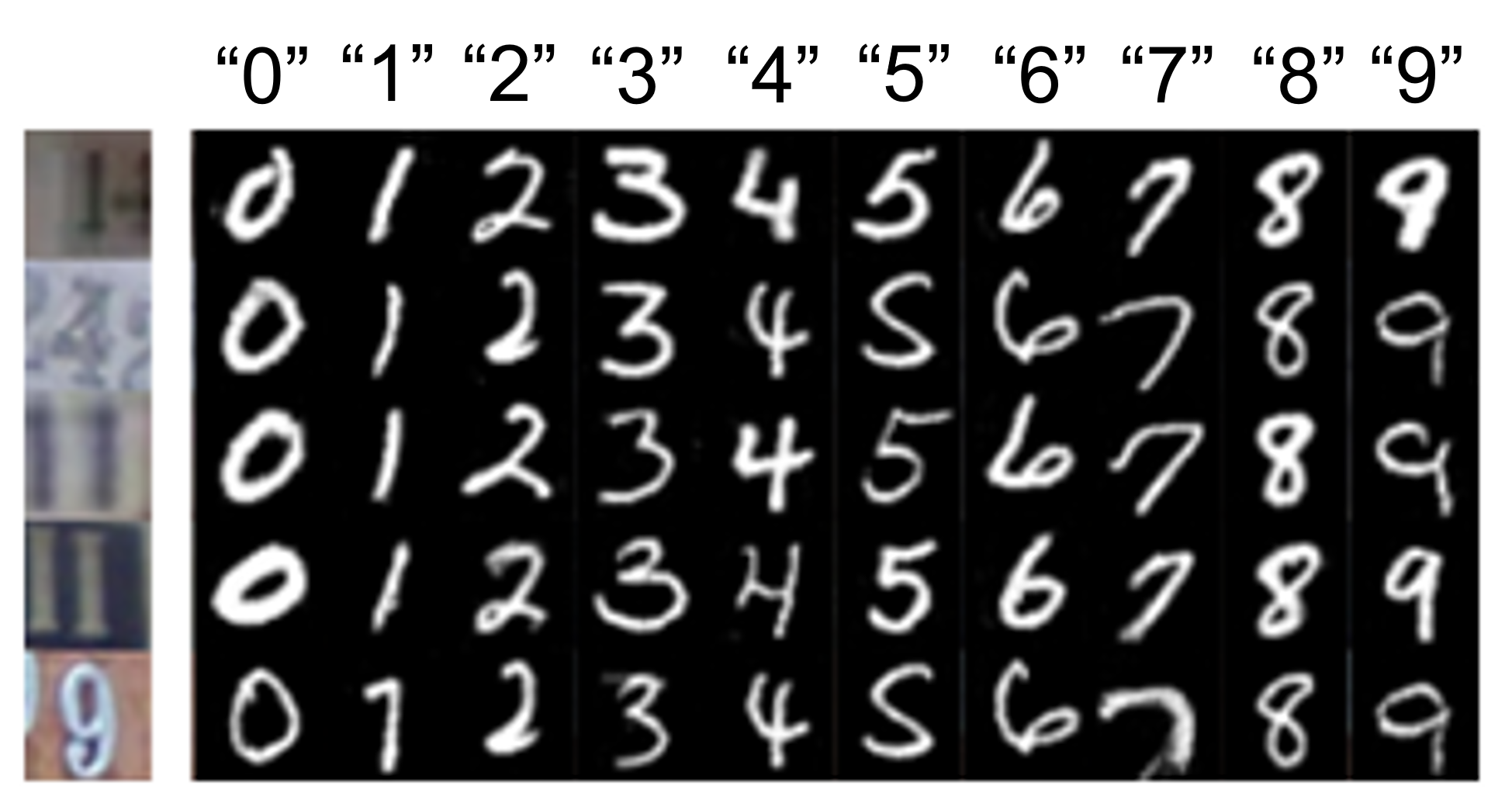} 
	\caption{Label-domination problem.
	The generator receives images from the leftmost column, paired with each label from ``0'' to ``9''. In the right part, the image in the $i$-th row and $j$-th column is the corresponding generated image of the $i$-th input image and label `$j$'. The output images are dominated by the input labels, showing no relationship with input images. More experiments demonstrating the existence of the label-domination problem can be seen in Appendix~C.} 
	\label{fig:label-domination}
\end{figure}

While the preliminary method resolves the label-inconsistency problem, we argue that this approach meets a major obstacle, referred to as the \emph{label-domination problem}, where the output of the conditional generator $G_{S\rightarrow T}(x, y)$ is dominated by the input label $y$ and the information of the source domain data $x$ is ignored.
We empirically identify the existence of the label-domination problem by constructing a series of experiments on $\mathcal{S}\rightarrow \mathcal{M}$. 
As shown in Fig.~\ref{fig:label-domination}, the left images and different labels are input to the conditional generator and the right-side images are the corresponding output. It shows that no matter what the input image is, the output will always follow the input label and show no correlation with the input images.
Refer to Appendix~C for ablation experiments that eliminate the effect of other factors including the network architectures, cycle consistency loss, and Spectral Normalization.

Intrinsically, the conditional discriminator in the target domain cannot avoid such a trivial solution.
That is because the generator can simply memorize some prototypes in each class and only generate the corresponding prototype according to the input label to cheat the discriminator. Such a trivial solution ignores the information of the source image and thereby would result in a poor adaptation.
% Furthermore, when the label-domination problem occurs, intuitively, the adaptation procedure will not benefit the training of the classification model. That is because the generator cannot transfer additional labeled data from the source domain, suggesting the poor usage of source image information.

As quantitatively shown in row 3 of Table.~\ref{tab:ablation}, this preliminary approach has roughly the same classification accuracy as semi-supervised learning, which validates our intuition.

\section{Method}
\label{sec:method}

\begin{figure*}[t]
    \centering
    \includegraphics[width=17cm]{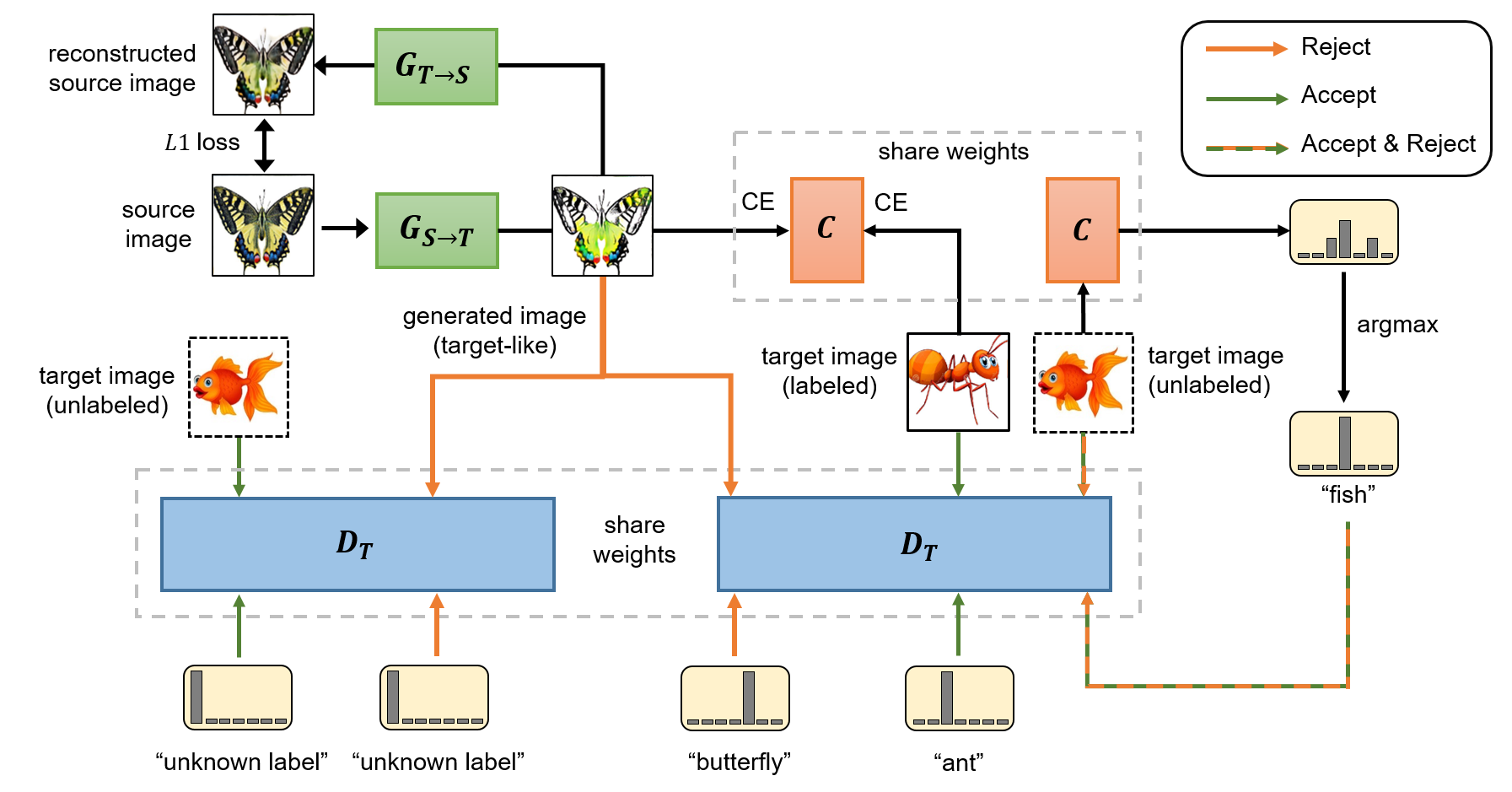}
    \caption{An overview of Relaxed cGAN. The discriminator takes image-label pairs as input while the generator takes source images as the only input, resolving both label inconsistency and label domination problem (see Sec.~\ref{subsec:relaxed-cgan}). The classifier is jointly trained with the generator and discriminator, ensuring a provable equilibrium (see Sec.\ref{subsec:theoretical-equilibrium}). Additionally, the discriminator accepts all real images in the target domain to leverage unlabeled data without changing the equilibrium (the left $D_T$, see Sec.~\ref{subsec:additional-enhancement}). Note that the discriminator randomly accepts the image-label pairs from the classifier to augment the labeled sample pool (see Sec.~\ref{subsec:experiment-setup}).
    % For discriminator, we employ two types of losses: (1) \textbf{Conditional loss} (the right $D_T$). The target image-label pairs are accepted and generated image-label pairs are rejected. We also use classifier $C$ to produce pseudo labels for unlabeled target images, and partially accept them to augment the positive samples. (2) \textbf{Marginal loss} (the left $D_T$). To make full use of unlabeled data, we feed them ``unknown label" and let discriminator treat them as positive samples while reject generated images with ``unknown label". Note that unlike conditional GAN~\cite{mirza2014conditional}, the generator does not accept label so that it shall infer the semantic of input image to fool discriminator, which alleviates label domination problem (see Sec.~\ref{sec:unconditional-generator}). The generated images and target labeled images are used for updating classifier using cross entropy loss (denoted by CE in the chart). 
    }
    \label{fig:flow-chart}
\end{figure*}

We first present the proposed Relaxed cGAN framework to solve the label-domination problem in Sec.~\ref{subsec:relaxed-cgan}. We then analyze it in terms of the theoretical equilibrium and empirical convergence in Sec.~\ref{subsec:theoretical-equilibrium}. Finally, we present additional regularization to enhance the Relaxed cGAN framework for SSDA in Sec.~\ref{subsec:additional-enhancement}. 

\subsection{Relaxed cGAN for a Provable Equilibrium}
\label{subsec:relaxed-cgan}

% \begin{figure*}[t]
% 	\centering
% 	\includegraphics[width=15cm]{figures/flowchart_GAN.png}
% 	\caption{An overview of Relaxed cGAN. For discriminator, we employ two types of losses: (1) \textbf{Conditional loss} (the right $D_T$). The target image-label pairs are accepted and generated image-label pairs are rejected. We also use classifier $C$ to produce pseudo labels for unlabeled target images and partially accept them to augment the positive samples. (2) \textbf{Marginal loss} (the left $D_T$). To make full use of unlabeled data, we feed them ``unknown label" and let discriminator treat them as positive samples while reject generated images with ``unknown label". Note that unlike conditional GAN~\cite{mirza2014conditional}, the generator does not accept label so that it shall infer the semantic of the input image to fool discriminator, which alleviates label domination problem (see Sec.~\ref{sec:unconditional-generator}). The generated images and target labeled images are used for updating the classifier using cross-entropy loss (denoted by CE in the chart). }
% 	\label{fig:flow-chart}
% \end{figure*}

The label-domination problem directly motivates our design of Relaxed cGAN.
Our key intuition is that the source-domain image itself contains semantic information and therefore the label is not necessary to be included in the generator.
Specifically, different from the preliminary approach, we feed images \textit{without} labels to our generator for adaptation. Then we construct the image-label pairs with the generated images and the labels inherited from the ground truth of the input images.
Such image-label pairs are used to fool the discriminator. By doing this, the generator is enforced to learn useful features from the raw images and generate meaningful samples in the target domain. 
Thus, the corresponding losses become
\begin{equation}
\begin{aligned}
\mathcal{L}_{GAN} &= \mathbb{E}_{(x,y) \sim {p}_{t}(x,y)}[\log (D_{T}(x,y))]  \\
&+ \alpha\mathbb{E}_{(x,y) \sim {p}_{s}(x,y)}[\log(1- D_{T}(G_{S\rightarrow T}(x), y))] \\
&+ (1-\alpha)\mathbb{E}_{x \sim {p}_{u}(x)}[\log(1-D_{T}(x, C(x)))],\label{eqn:triple-gan-loss}
% &+\mathbb{E}_{(x,y) \sim {p}_{s}(x,y)}[\log (D_{S}(x,y))]  \\
% &+ \mathbb{E}_{(x,y) \sim {p}_{t}(x,y)}[\log(1- D_{S}(G_{T\rightarrow S}(x), y))] \\
\end{aligned}
\end{equation}
and the cycle loss is 
\begin{equation*}
\begin{aligned}
\mathcal{L}_{cycle} &= \mathbb{E}_{(x,y) \sim {p}_{s}(x,y)}[|G_{T\rightarrow S}(G_{S\rightarrow T}(x))-x|] \\
&+ \mathbb{E}_{(x,y) \sim {p}_{t}(x,y)}[|G_{S\rightarrow T}(G_{T\rightarrow S}(x))-x)|], 
\end{aligned}
\end{equation*}
% \begin{equation*}
% \begin{aligned}
% \mathcal{L}_{C} &= \frac{1}{K}\mathbb{E}_{x \sim p_u(x)}[\sum_{i=1}^{K}-C(i|x)\log C(i|x)]. \\
% &+ \mathbb{E}_{(x,y) \sim p_s(x,y)}[-\log(C(y|G_{S \rightarrow T}(x)))]
% \end{aligned}
% \end{equation*}
where $C(x)$ is the classifier's predicted class for image $x$. The third term and the loss weight of $(1-\alpha)$ in $\mathcal{L}_{GAN}$ is the adversarial training between the classifier and the discriminator. For simplicity we set $\alpha=\frac{1}{2}$ following ~\cite{chongxuan2017triple}. They are used to guarantee our method for a provable Nash equilibrium, referring to the next section. Except for these factors, the main difference from the losses in the preliminary method is that the inputs to both generators are the images only, while the discriminator still receives the image-label pairs as input.
% Additionally, we enforce the discriminator to discriminates the data label pairs from classifier, and we claim that such design would ensure a provable equilibrium that $p_t(x,y)=p_g(x,y)=p_c(x,y)$ (see details in Appendix A).

% To further increase the model's ability of making use of unlabeled data, we adopt a marginal loss, where the discriminator accepts data only instead of data label pair. We use extra class (the $(K+1)-th$ class) as input label:

% \begin{equation*}
% \begin{aligned}
% \mathcal{L}_{marg} &= \mathbb{E}_{(x,y) \sim {p}_{t}(x,y)}[\log (D_{T}(x,K+1))]  \\
% &- \mathbb{E}_{(x,y) \sim {p}_{s}(x,y)}[\log(1- D_{T}(G_{S\rightarrow T}(x), K+1))], \\
% &+\mathbb{E}_{(x,y) \sim {p}_{s}(x,y)}[\log (D_{S}(x,K+1))]  \\
% &- \mathbb{E}_{(x,y) \sim {p}_{t}(x,y)}[\log(1- D_{S}(G_{T\rightarrow S}(x), K+1))], \\
% \end{aligned}
% \end{equation*}

\subsection{Theoretical and Empirical Analysis}
\label{subsec:theoretical-equilibrium}

We now analyze Relaxed cGAN in terms of theoretical equilibrium, practical convergence, and image transfer results.

\textbf{Theoretical Equilibrium.}
We prove that the minimax game defined by Relaxed cGAN has a unique global equilibrium (under the non-parametric assumption~\cite{goodfellow2014generative}) that the generated joint distribution, the classifier distribution, and the real target distribution are the same.

Formally, let $p_g(x,y)$ denote the joint distribution characterized by the image-label pairs of generated images and their original labels, $p_c(x,y)$ denote the distribution of image-label pairs constructed from the target domain images and the corresponding predictions by the classifier, and $p_t(x,y)$ denote the target domain joint distribution as mentioned in the Sec.~\ref{subsec:setting-framework} settings. We have the following theorem to show that our design theoretically has a desirable equilibrium. The detailed proof is given in Appendix~A.

\begin{theorem}
Under the non-parametric assumption, the equilibrium of $\tilde{\mathcal{L}}(C, G_{S \rightarrow T}, D_T)$ is achieved if and only if $p_t(x,y)=p_g(x,y)=p_c(x,y)$, where $\tilde{\mathcal{L}}$ is defined as
$$
\tilde{\mathcal{L}}(C, G_{S \rightarrow T}, D_T) = \mathcal{L}_{GAN} + \mathcal{L}_{C} + \mathcal{L}_{cycle}.
$$ 
\end{theorem}

%  Secondly we adopt a visualization tool called Path Angle to empirically show the superiority of our framework under the practical scenery when optimizing the GAN loss.
% We demonstrate the rationality of our design in this section. 
% GAN NE theoretically.
% But negative results -> maybe under sgd we cannot get NE. Path Angle is a visualization method ...

% General structure
% start from cGAN.
% label domination: Existence, Solution, Explanation(proof, path-angle).

% We visualize training dynamics of rcGAN and rcTGAN with label in G to show the superiority of our design.
% Path angle is a powerful method to show ... based on vector field.
% We explore the scenario of S2M with network in fig 6.
% cos_sim:..., grad_total_norm:...
% RcGAN shows change of direction & smaller grad norm -> RcGAN is more close to LSSP. RcGAN with y in G cos_sim=0, grad_norm continues to decrease, it rotate and is hard to converge to LSSP. label domination existing in such design neglecting input image and only output several images, can be regarded as intra-class mode collapse -> rotating.  

\begin{figure}[t]
    \centering
    \subfigure[Relaxed cGAN]{\includegraphics[width=4.2cm]{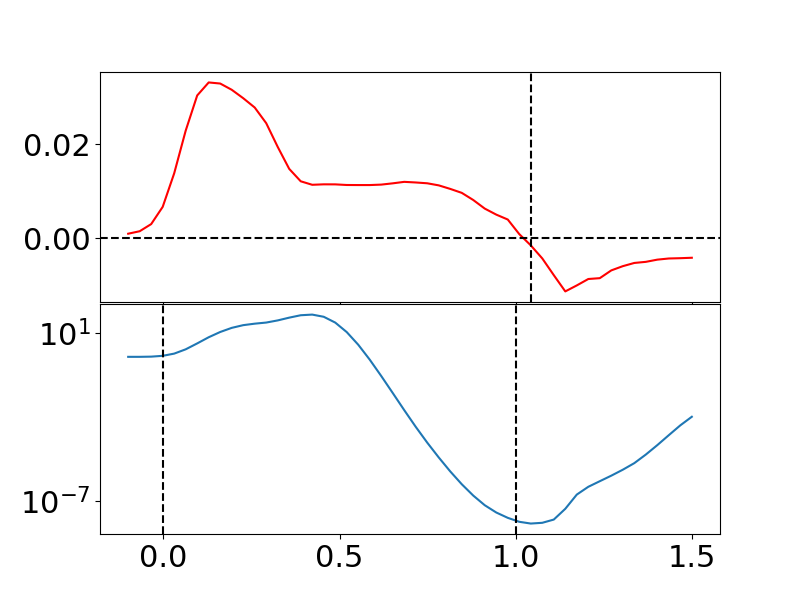}}\subfigure[Preliminary cGAN]{\includegraphics[width=4.2cm]{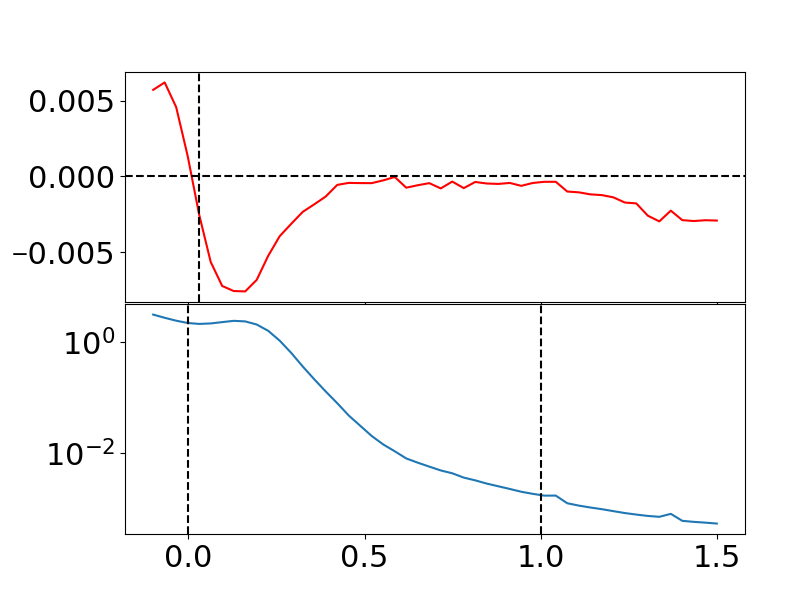}} \hspace{7mm}
    
    \subfigure[Attraction Field]{\includegraphics[width = 3.7cm]{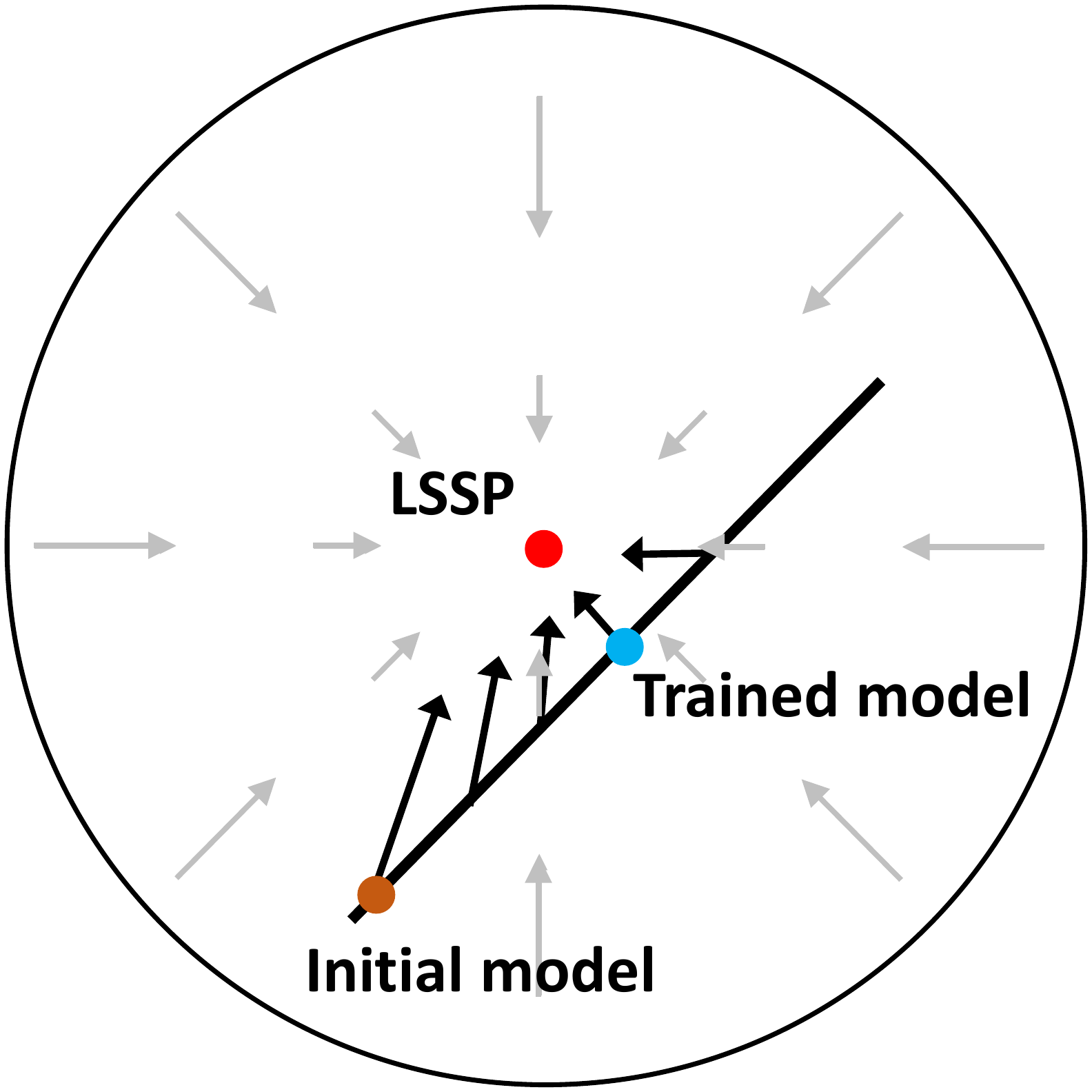}}\hspace{7mm}
    \subfigure[Rotation Field]{\includegraphics[width = 3.7cm]{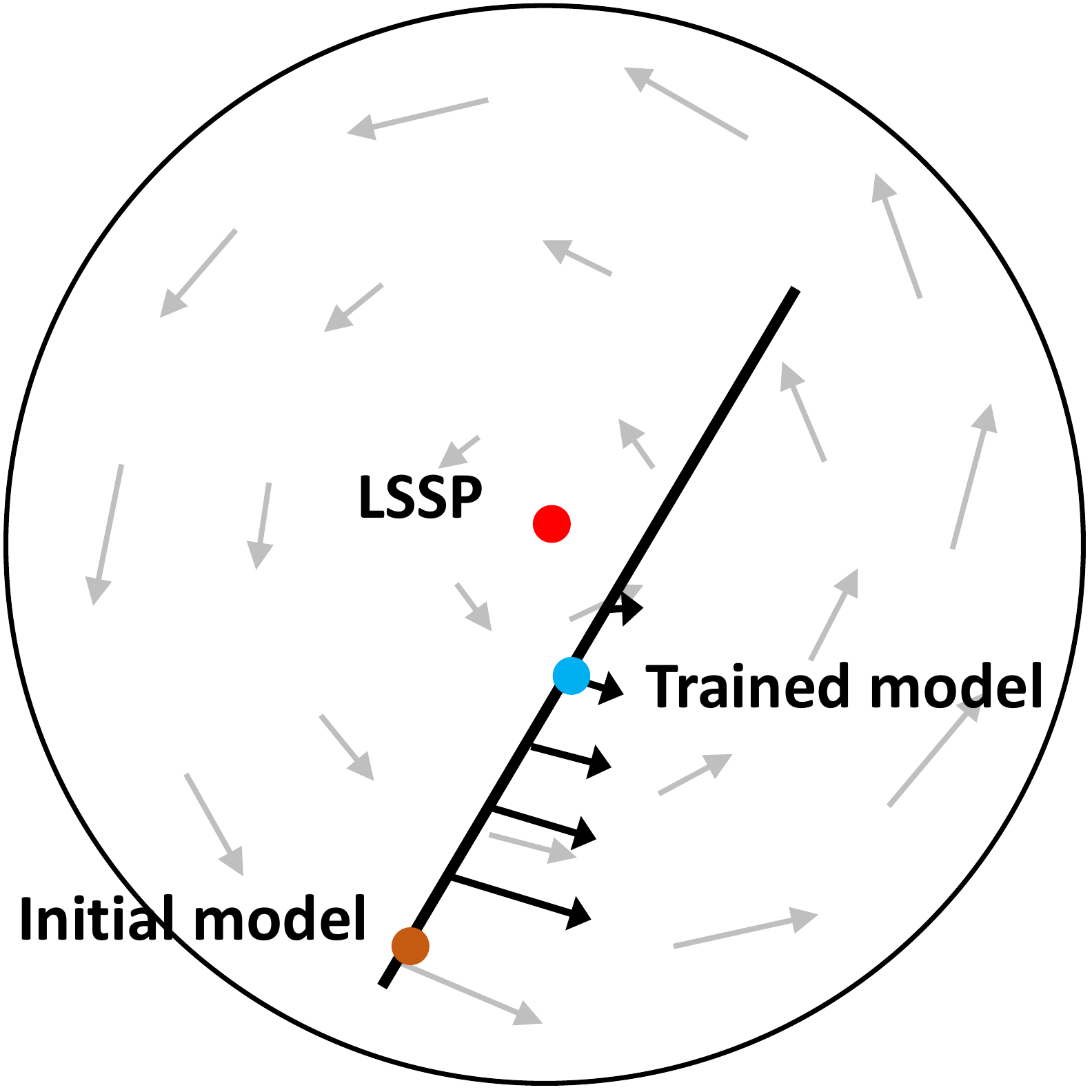}}

    \caption{Visualization of the practical convergence of Relaxed cGAN and Preliminary cGAN. (a) and (b) visualize their vector field statistics. The upper red line is the cosine angle between the gradient vector and the interpolation path. The lower blue line is the corresponding gradient norm. The X-axis means the interpolation point. (c) and (d) are the game vector field imitated for Relaxed cGAN and Preliminary cGAN respectively. Black arrows show the gradient direction at different linear interpolations between the initial model and the final trained model. The gradients of Relaxed cGAN point to the LSSP while in Preliminary cGAN they are perpendicular to the interpolation line. It shows that Relaxed cGAN performs better than the Preliminary cGAN under practical training.}
    % (b) The cosine and gradient norm for Conditional GAN. (c) The imitated game vector field for Relaxed cGAN. The \zhili{red point} is LSSP and the blue line is the interpolating line. Black arrows correspond to the directions of the vector field at different linear interpolations between the initial and final points. (d) The imitated vector field for Conditional GAN. }
    \label{fig:gan-training}
\end{figure}

\textbf{Practical Convergence.}
Despite the theoretical proof of the Nash equilibrium, it is insufficient to guarantee our model achieving such equilibrium under the practical optimization with stochastic gradient decent~\cite{mescheder2017numerics,balduzzi2018mechanics,gidel2019negative}. Therefore, we adopt a tool called Path Angle~\cite{berard2019closer} to verify the effectiveness of our model's practical convergence property.

In practice, studying the game vector field (i.e., the concatenation of both networks' gradient) around the local stable stationary point (LSSP), which is a necessary condition for Nash equilibrium, is an alternative point of view that can provide better insights of the practical convergence~\cite{mescheder2017numerics,gidel2018variational}. Compared to Nash equilibrium which corresponds to the static stability, LSSP captures the dynamic stability of the training process. 
% Path Angle~\cite{berard2019closer} visualizes the vector field around the LSSP by interpolating between the initial training parameters and the final model (correspond to the blue line and yellow line in Fig.~\ref{fig:gan-training}(a) and (b)), and calculate the gradient norm and the angle between the gradient vector and the interpolation line. We refer readers to the original paper for more details.
Path Angle provides insights about the convergence property around the LSSP. To calculate it, we first linearly interpolate models between the initial parameters and the final parameters, then compute gradients for these models. The convergence characteristic around the LSSP can be analyzed by (1) the gradient norm and (2) the angle between the gradient vector and the interpolation line. We refer readers to the original paper for more details.

As shown in Fig.~\ref{fig:gan-training}(a), relaxed cGAN shows a change of direction and smaller gradient norm when the training is close to the convergence (represented as point 1.0 of x-axis). Such a pattern shows an attraction field referring Fig.~\ref{fig:gan-training}(c) where the gradient direction always points to the LSSP and the norm decreases as being close to it. The mode indicates that during training, relaxed cGAN can converge to a point that is close to LSSP and thus can get excellent performance.

For comparison, we also visualize the training dynamics of the preliminary cGAN framework introduced in Sec.~\ref{sec:preliminary} . As shown in Fig.~\ref{fig:gan-training}(b), the preliminary method always has the cosine value close to 0, showing rotation phenomena. Besides, the gradient norm is relatively large and continues to decrease, which means that the preliminary method is still far from the LSSP but the model cannot reach there because of rotation. Such poor practical convergence property may offer a possible explanation of why even though the preliminary method have the same Nash equilibrium as Relaxed cGAN does~\cite{goodfellow2014generative}, it encounters the label-domination problem and gives poor results for classification — the model always outputs different prototypes during training just like rotation but cannot converge to an LSSP.

\begin{figure}[htbp]
    \centering

    \subfigure[Digit: SVHN to MNIST ]{\includegraphics[height = 2cm]{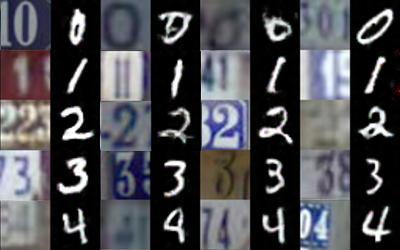}}
    %\subfigure[DomainNet: Real to Clipart]{\includegraphics[height = 2cm]{figures/Fig5a.png}}
    \subfigure[DomainNet: Clipart to Sketch]{\includegraphics[height = 2cm]{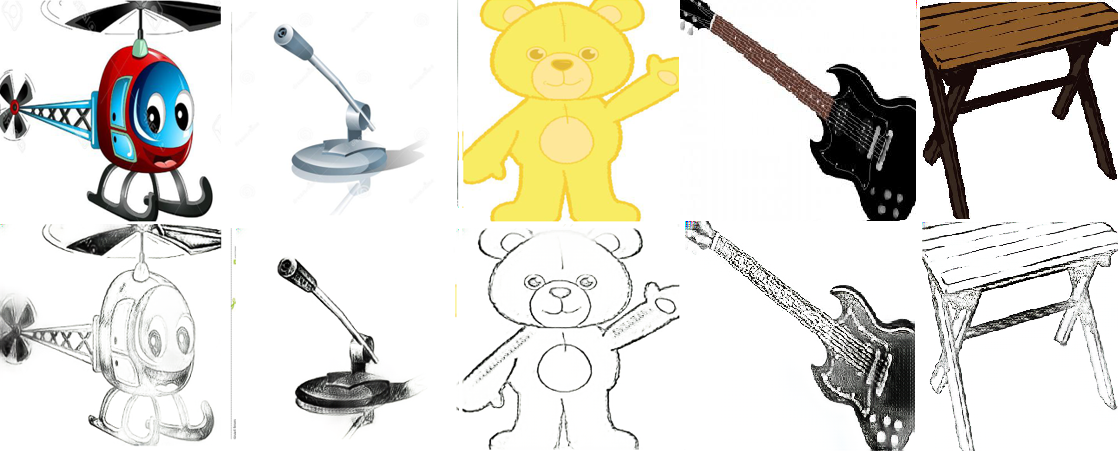}}
    \caption{The generation results of Relaxed cGAN. For the digit experiment, the odd column is the input image and the even column is the generated image. For DomainNet experiment, the upper row is the input image and the lower row is the generated image. As we can see, the generated images are of high quality and their labels follow the ground truth of input labels.}
    \label{fig:rcgan-generation}
    \vspace{-0.2cm}
\end{figure}

\textbf{Image Transfer Results.}
We further validate our design in terms of image transformation. As shown in Fig.~\ref{fig:rcgan-generation}, the generated samples are of high quality.

% \begin{figure}[thb]
%     \centering
%     % \subfigure[CycleGAN]{\includegraphics[height = 2.5cm]{figures/Fig2b_n.png}}
%     % \subfigure[Relaxed cGAN]{\includegraphics[height = 2.5cm]{figures/Fig2a_n.png}}
%     \includegraphics[height = 2.5cm]{figures/Fig2a_n.png}
%     \caption{The generation result of Relaxed cGAN on S $\rightarrow$ M. As we can see, the generated images are of high quality and their labels follow the ground truth of input labels}
%     \label{fig:rcgan-generation-s2m}
% \end{figure}

% For classification, Relaxed cGAN can outperform the naive approach described above, as shown in rows 4 and 5 of Table~\ref{tab:ablation}. These two experiments directly show the effectiveness of our framework for solving the label domination problem and thus benefit the adaptation procedure.

% We take the assumption that under practical training, the preliminary method cannot reach a good LSSP. We regard label domination as a consequence of rotation phenomena because the model always outputs different prototypes during training but cannot converge to a LSSP.

% Actually the preliminary method can also have the same nash equilibrium as our model does, however, it encounters the label-domination problem and gives poor results for classification. We take the assumption that under practical training, our design can get a better local nash equilibrium than the preliminary method. We verify our assumption through a powerful tool called Path Angle, to visualize the training dynamics of our model and the preliminary method. 

% Though the proposed Relaxed cGAN has the same theoretical convergence point as the preliminary method, we empirically show that the Relaxed cGAN has a better practical convergence property with a tool named path angle \ref{}. As shown in Fig.~\ref{fig:gan training}

\subsection{Additional Regularization}
\label{subsec:additional-enhancement}

Apart from the label-domination problem, it is crucial to leverage the large amount of unlabeled data in the target domain to improve the results in SSDA. 
To this end, we add a marginal loss such that the discriminator accepts all the real images in the target domain (including labeled data and unlabeled data). 
In particular, we feed an extra $(K+1)$-th class to the discriminator (same discriminator as before), where $K$ is the number of classes of the data. 
All the target images belong to the $(K+1)$-th class. 
The marginal loss function for $D$ is defined as:
% \begin{equation}
% \begin{split}
% \mathcal{L}_{marg-D} &= -\mathbb{E}_{x \sim {p}_{u}(x)}[\log (D_{T}(x, K+1))] \\
% & - \mathbb{E}_{x \sim {p}_{s}(x)}[\log (1-D_{T}(G_{S\rightarrow T}(x), K+1))],
% \end{split}
% \label{eq:marginal-loss}
% \end{equation}
\begin{equation*}
\begin{aligned}
\mathcal{L}_{marg} &= \mathbb{E}_{x \sim {p}_{t}(x)}[\log (D_{T}(x,K+1))]  \\
&+ \mathbb{E}_{x \sim {p}_{u}(x)}[\log (D_{T}(x,K+1))] \\
&+ \mathbb{E}_{x \sim {p}_{s}(x)}[\log(1- D_{T}(G_{S\rightarrow T}(x), K+1))], \\
% &+\mathbb{E}_{(x,y) \sim {p}_{s}(x,y)}[\log (D_{S}(x,K+1))]  \\
% &- \mathbb{E}_{(x,y) \sim {p}_{t}(x,y)}[\log(1- D_{S}(G_{T\rightarrow S}(x), K+1))], \\
\end{aligned}
\end{equation*}
The first two items are positive samples for the discriminator from the labeled and unlabeled target data. The third item represents the negative samples as they come from the generator.

Intuitively, the marginal loss can avoid that the discriminator only memorizes the empirical distribution of labeled data.
Further, we prove that the marginal loss would not change the equilibrium of the three networks (see Appendix~A for more details), and we verify the effectiveness of the marginal loss in both image classification (see Table.~\ref{tab:ablation}) and image transfer (see Appendix~D).

% in Eqn.~\eqref{eq:pseudo-loss} and Eqn.~\eqref{eq:marginal-loss} in terms of both image transfer and classification.
% As shown in Fig.~\ref{fig:pseudo-marginal-loss}, Relaxed cGAN without marginal loss generates images very similar to the labeled data in the target domain, while standard Relaxed cGAN generates images of greater diversity. 
% Such diverse samples can provide useful learning signals to the classifier and lead to more accurate predictions.
% See results in rows 3 and 5 in Table~\ref{tab:ablation}. 

Therefore, the overall loss function would be 
\begin{equation*}
\begin{aligned}
\mathcal{L} = \lambda_{GAN}\mathcal{L}_{GAN} + \lambda_{cycle}\mathcal{L}_{cycle} + \lambda_{C}\mathcal{L}_{C} + \lambda_{marg}\mathcal{L}_{marg}. 
\end{aligned}
\end{equation*}

\section{Experiments}
\label{sec:experiment}

\begin{table*}[ht]
  \centering
  \footnotesize{\scalebox{1}{\setlength{\tabcolsep}{1.8mm}{
    \begin{tabular}{cccccccc}
    \toprule
    \multicolumn{1}{r}{} & \multicolumn{1}{r}{} & \multicolumn{4}{c}{ pixel-level } & \multicolumn{2}{c}{feature-level} \\
    \cmidrule(r){3-6} \cmidrule(r){7-8}
    \textbf{dataset} & \textbf{\# of label} & \textbf{Relaxed cGAN} & \textbf{cycleGAN~\cite{zhu2017unpaired}} & \textbf{CyCADA~\cite{hoffman2017cycada}} & \textbf{ACAL~\cite{hosseini2018augmented}} & \textbf{AVDA~\cite{perez2019matching}} & \textbf{F-CADA~\cite{zou2019consensus}} \\
    & & & +$\mathcal{L}_C$+$\mathcal{L}_{ENT}$ & +$\mathcal{L}_C$ & +$\mathcal{L}_{ENT}$ & &(best) \\ 
    \midrule
    \multirow{7}{*}{ $\mathcal{S}\rightarrow \mathcal{M}$ } & 1     & \textbf{95.9±2.8} & 70.2±6.8  & 90.6±0.1 & 80.5±0.8  & 95.1±0.5 & 94.8 \\
      & 2     & \textbf{96.3±1.3} & 81.3±3.0  &  90.8±0.3     & 94.9±0.9  &    -   & 95.1 \\
      & 3     & \textbf{96.9±0.4} & 81.7±3.9  &  90.8±0.5     & 94.5±0.3  &    -   & 95.4 \\
      & 4     & \textbf{97.0±0.4} & 83.3±2.9  &  90.7±0.2     & 94.8±0.5  &    -   & 95.5 \\
      & 5     & \textbf{97.5±0.3} & 84.5±3.6  &  90.6±0.6     & 94.9±0.6  & 96.9±0.2 & 95.6 \\
      & 6     & \textbf{98.0±0.5} & 86.4±2.4  &  91.5±0.3     & 96.0±0.9  &    -   & 95.9 \\
      & 7     & \textbf{98.0±0.5} & 85.3±2.9  &  91.3±0.2     & 96.3±0.7  &    -   & 96.1 \\
    \midrule
    \multirow{7}{*}{$\mathcal{U}\rightarrow \mathcal{M}$} & 1     & \textbf{98.7±0.6} & 78.7±0.9  & 96.0±0.3      & 81.3±9.3      & 98.3±0.3 & 97.5 \\
      & 2     & \textbf{99.0±0.2} & 82.4±4.0  & 96.2±0.2      & 93.2±2.9      & -      & 97.8 \\
      & 3     & \textbf{99.0±0.2} & 86.3±5.1  & 96.1±0.1      & 89.2±6.8      & -      & 98.1 \\
      & 4     & \textbf{99.1±0.1} & 82.5±2.3  & 96.5±0.1      & 92.4±3.9      & -      & 98.4 \\
      & 5     & \textbf{99.1±0.1} & 83.9±3.5  & 96.3±0.2      & 94.4±1.1      & 98.6±0.03 & 98.6 \\
      & 6     & \textbf{99.2±0.1} & 84.5±3.9  & 96.5±0.2      & 95.3±0.8      & -      & 98.8 \\
      & 7     & \textbf{99.2±0.1} & 82.9±2.9  & 96.3±0.2      & 95.6±0.6      & -      & 98.9 \\
    \midrule
    \multirow{7}{*}{$\mathcal{M}\rightarrow \mathcal{U}$} & 1     & 97.9±0.5 & 81.7±1.3 & 94.2±0.4      &  68.3±12.9     & \textbf{98.4±0.3} & 97.2 \\
    & 2     & \textbf{98.0±0.2} & 84.7±2.2 & 94.0±0.3      &  86.1±2.9     &   -    & 97.5 \\
      & 3     & \textbf{98.0±0.3} & 85.0±4.7 & 94.4±0.4      & 82.0±1.6      & -      & 97.9 \\
      & 4     & \textbf{98.2±0.3} & 81.4±3.1 & 95.0±0.3      & 84.8±8.7      & -      & 98.1 \\
      & 5     & 98.2±0.1 & 85.5±1.2      &  95.0±0.1     &  88.2±2.7     & \textbf{98.5±0.2} & 98.3 \\
      & 6     & \textbf{98.4±0.1} & 86.4±5.2      &  94.5±0.5     &  83.0±1.6     &   -    & 98.4 \\
      & 7     & \textbf{98.2±0.1} & 82.9±3.2      &  95.1±0.2     &  90.2±2.7     &  -     & \textbf{98.6} \\
    \bottomrule
    \end{tabular}}}}%
  \caption{Digit High-resource SSDA. CyCADA~\cite{hoffman2017cycada} and ACAL~\cite{hosseini2018augmented} are two pixel-level methods proposed for Unsupervised Domain Adaptation and Supervised Domain Adaptation respectively. We extend their method for SSDA by adding the supervised loss $\mathcal{L}_C$ and the semi-supervised loss $\mathcal{L}_{ENT}$ for a fair comparison. AVDA~\cite{perez2019matching} and F-CADA~\cite{zou2019consensus} are two feature-level methods. All experiments are performed 3 times and we report both the average accuracy and standard deviation. Our method outperforms all baselines on $\mathcal{S}\rightarrow \mathcal{M}$ and $\mathcal{U}\rightarrow \mathcal{M}$, and achieve comparable results on $\mathcal{M}\rightarrow \mathcal{U}$.}
  \label{tab:high-resource-ssda}%
\end{table*}%

\begin{table*}[ht]
%~\emph{Note:} for each column, we have $10$\% labeled sample. For instance, the first column contains 1 labeled target sample and 9 unlabeled target samples per class.}
\centering
\footnotesize{\scalebox{1}{\setlength{\tabcolsep}{1.8mm}{
\begin{tabular}{lccc ccc ccc ccc}
\toprule
 & \multicolumn{3}{c}{ $\mathcal{S}\rightarrow \mathcal{M}$} & \multicolumn{3}{c}{  $\mathcal{M}\rightarrow \mathcal{U}$} &
   \multicolumn{3}{c}{$\mathcal{U}\rightarrow \mathcal{M}$} &
   \multicolumn{3}{c}{$\mathcal{S}\rightarrow \mathcal{U}$}\\ 
   \cmidrule(r){2-4} \cmidrule(r){5-7} \cmidrule(r){8-10} \cmidrule(r){11-13}
\# of labels per class &  $5$ & $10$ &  $600$ & $5$ & $10$ & $72$ &
   $5$ & $10$ & $600$ &
   $5$ &  $10$ & $72$ \\
 \# of data per class&  $50$ & $100$ &  $6000$ & $50$ & $100$ & $729$ &
   $50$ & $100$ & $6000$ &
   $50$ &  $100$ & $729$ \\
\midrule
% \small target-only      & \small 89.8 & \small 93.2 & \small 98.9 & \small 85.9 & \small 91.8 & \small 96.7 & \small 89.8 & \small 93.0 & \small 99.0 & \small 85.1 & \small 91.5 & \small 96.4 \\
% \small $n=200$ & \small  & \small  & \small  & \small  & \small  \\
cycleGAN~\cite{zhu2017unpaired} &  68.3 & 81.3 &  96.0 &  94.5 & 95.8 &  95.9 &  92.4 & 92.8 &  97.0 & 66.7 &  77.5 &  90.6 \\
 ACAL~\cite{hosseini2018augmented}             &  87.2 &  91.8 & 99.4 &  94.2 &  96.0 &  95.7 &  96.8 &  96.9 &  98.5 &  86.2 &  89.0 &  93.2 \\
 Relaxed cGAN    &  \textbf{96.9} &  \textbf{97.7} &  \textbf{99.6} &  \textbf{97.1} &  \textbf{97.5} &  \textbf{98.1} &  \textbf{96.9} &  \textbf{97.1} &  \textbf{99.6} &  \textbf{94.6} &  \textbf{96.4} &  \textbf{98.2}\\
\bottomrule
\end{tabular}}}}
\caption{Digit Low-resource SSDA. We follow the settings of the pixel-level method ACAL where unlabeled data in the target domain are also limited. Relaxed cGAN achieves SOTA in all settings.}
\label{tab:semi-supervised domain adaptation}
\end{table*}

% Setup
\subsection{Setup}
\label{subsec:experiment-setup}
% In this section, we present the detail of our experiment setting and results. 
%In this section, we evaluate the proposed model under two widely-used  SSDA settings, including (1) %\textbf{high-resource} SSDA where one has a few labeled data together with sufficient unlabeled data, 
We first evaluate the proposed Relaxed cGAN in two practical SSDA settings on Digit adaptation benchmarks, including (1) \textbf{High-resource} SSDA, where we have full access to the target dataset's unlabeled data and a few labeled data. (2) \textbf{Low-resource} SSDA, where the number of unlabeled data in the target domain is also limited. 
Secondly, we verify our approach on several large-scale image datasets, namely, DomainNet~\cite{peng2019moment} and Office-Home\cite{venkateswara2017deep}.
% In these two settings, our approach outperforms recently proposed SSDA methods, showing the ability to utilize source domain data and target domain unlabeled data effectively.

\paragraph{Datasets.}
% Datasets
%     digit(copy)
%     domainNet(follow MME)
% We use three benchmark digit datasets including MNIST, USPS, and Street View House Numbers (SVHN). 
We first perform experiments on three benchmark \textbf{Digit} datasets, including MNIST~\cite{lecun1998gradient}, USPS~\cite{denker1989neural} and SVHN~\cite{netzer2011reading}, all of which are 10-class classification tasks.
Both MNIST and USPS are datasets consisting of grayscale images of handwritten digits. 
% MNIST contains 70,000 images, with 60,000 training samples and 10,000 test samples. USPS is a relatively small dataset, with 7,291 training samples and 2,007 test samples. 
SVHN is a more diverse house number dataset which is obtained by cropping Google Street View images.
% It consists of 73,257 training samples and 26,032 test samples.
For \textbf{DomainNet}, we follow the experiment settings of \cite{saito2019semi} and choose 4 domains: Real(R), Clipart(C), Sketch(S) and Painting(P), with 126 shared classes across domains.
\textbf{Office-Home} is another standard benchmark for domain adaptation, containing 4 domains (Real, Clipart, Art, and Product).

\paragraph{Implementation Details.}
% Implementation details
%     Preprocessing
%         digit(copy)
%         domainNet(follow MME)
%     Network details(no detail, see appendix)
%         digit
%         domainNet(follow MME, alexnet, GD)
%     Training techniques.
%         domainNet - source sup loss
%         domainNet - sn and hinge loss
\textbf{(1) Preprocessing.} 
% since it's a widely used protocal, we don't mention it in the paper.
All images are normalized with a mean of 0.5 and a standard deviation of 0.5. 
For digit tasks, images are resized to $32\times32$. We duplicate MNIST and USPS image channels 3 times to match the channel size of SVHN. 
For DomainNet, images are resized to $256\times256$, according to our design of the generator.
\textbf{(2) Network Details.} 
For adaptation between USPS and MNIST, we employ a simple architecture. For transfer tasks from SVHN to MNIST and SVHN to USPS, we adopt a slightly complex architecture.
For DomainNet, we use imagenet-pretrained Alexnet~\cite{krizhevsky2012imagenet} as the classifier. 
The base architecture of the generator and the discriminator mainly follows the default design of cycleGAN, except conditional input and Spectral Normalization. 
See more details in Appendix B.
\textbf{(3) Training Techniques.}
% We adopt the training techniques of Triple-GAN. Entropy minimization is applied to unlabeled data to extract more discriminative features. To augment the labeled samples for $D$, we generate pseudo labels with $C$ for randomly selected images and feed the label-data pairs as positive samples to $D$. By doing this, $D$ would access more diverse data instead of the limited labeled samples in the target domain. During the training, we also use the generated data to train $C$, serving as additional data augmentation.
To further make use of unlabeled data, we apply widely adopted entropy minimization loss~\cite{grandvalet2005semi} denoted as $\mathcal{L}_{ENT}$.
For DomainNet and Office-Home, we adopt source data supervision as a regularization. To avoid that the discriminator overfits on labeled images, we generate pseudo labels using the classifier $C$ for randomly selected unlabeled images and feed the image-label pairs as positive samples to $D_T$ for training. Corresponding loss is defined as follows:
\begin{equation*}
\mathcal{L}_{pseudo} = \mathbb{E}_{x \sim {p}_{u}(x)}[\log (D_{T}(x, C(x)))].
\end{equation*}
Note that this loss performs conversely against the third item in Eqn.~\eqref{eqn:triple-gan-loss} which may introduce some biases to the target distribution, resulting in a mixture distribution of $p_c$ and $p_t$. However, since $p_c$ and $p_t$ are close, such bias may be neglected in practice.

\paragraph{Baselines.}
%
% Baselines:
% feature/pixel level diff.
%     (1) Digit high
%     (2) Digit low
%     (3) DomainNet
%
% Baselines
%     digit-baseline(copy)
%     domainNet - MME
% \zhili{related work should mention pixel-level}
We mainly divide our baselines into two categories: feature-level and pixel-level adaptation. 
\textbf{(1) Digit High-resource SSDA.} We compare our proposed method with two feature-level methods, F-CADA~\cite{zou2019consensus} and AVDA~\cite{perez2019matching}. Since few pixel-level methods explore the setting of SSDA, we compare with CyCADA~\cite{hoffman2017cycada} and ACAL~\cite{hosseini2018augmented} which are strong methods proposed for Unsupervised Domain Adaptation(UDA) and Supervised Domain Adaptation(SDA) respectively. For a fair comparison, We extend their methods by adding the supervised loss $\mathcal{L}_C$ and the semi-supervised loss $\mathcal{L}_{ENT}$, respectively.
%  F-CADA maps both the source and target domains' features into a shared space, and utilizes annotated data by minimizing information entropy loss. 
% AVDA aligns samples of different classes with corresponding Gaussian mixture components in an adversarial way.
% We also compare with two supervised domain adaptation (SDA) baselines for reference, including FADA~\cite{motiian2017few} and CCSA~\cite{motiian2017unified}.
\textbf{(2) Digit Low-resource SSDA.} It is an important setting proposed by ACAL~\cite{hosseini2018augmented} that few unlabeled data on the target domain are available. We also compare our model with their method. 
% ACAL adopts a classifier as a complementary discriminator so that the semantic information of generated images can be maintained.
The results of cycleGAN are also listed for reference.
\textbf{(3) DomainNet \& Office-Home.} MME~\cite{saito2019semi}, MetaMME~\cite{li2020online} and APE~\cite{kim2020attract} are 3 feature-level methods. A recent work called BiAT~\cite{jiangbidirectional} generates images by adversarial training, so we choose their work as our pixel-level baseline.

% state-of-the-art approach on DomainNet called MME~\cite{saito2019semi}, which adapts the model's prototype to target domain by iteratively minimaxing the conditional entropy of target domain unlabeled data.

\subsection{Results}
% Results
%     digit high-resource(copy)
%     digit low-resource(copy), unlabel exp
%     domainNet
%     ablation(be referenced from method)

% \subsubsection{Digit High-resource SSDA}
\textbf{Digit High-resource SSDA.}
In the Digit high-resource SSDA settings, we could access the whole training dataset of the target domain with partially labeled data. We adopt the same experimental settings as in  F-CADA~\cite{zou2019consensus} and AVDA~\cite{perez2019matching}, where the number of the labeled data in each class is set from 1 to 7. Following AVDA~\cite{perez2019matching}, we report both the averaged accuracy (with the standard deviation) over three runs as well as the best of them. 
As shown in Table~\ref{tab:high-resource-ssda}, both the average and the best results of Relaxed cGAN consistently outperform the SSDA baselines under $\mathcal{S}\rightarrow \mathcal{M}$ and $\mathcal{U}\rightarrow \mathcal{M}$, suggesting that Relaxed cGAN performs effective adaptation.

In $\mathcal{M}\rightarrow \mathcal{U}$, the number of unlabeled data is about ten percent of that in the former two settings, making the adaptation process harder. Nevertheless, Relaxed cGAN still achieves the best performance in five out of the seven settings and is competitive with the strong feature-level baselines in the other two settings.

% \subsubsection{Digit Low-resource SSDA}
\textbf{Digit Low-resource SSDA.}
In the low-resource SSDA settings, the number of unlabeled data in the target domain is also limited. We compare against the representative pixel-level baselines cycleGAN~\cite{zhu2017unpaired} and  ACAL~\cite{hosseini2018augmented}, following the same experiment settings in ACAL for fairness.
The number of available images in the target domain is set to 50, 100, and full per class, and the labeled samples take 10\% of the available target samples. We present the experiment results in Table \ref{tab:semi-supervised domain adaptation}. 
It can be seen that our method significantly outperforms cycleGAN and ACAL in all settings even though we do not apply any form of semantic consistency loss as mentioned in ACAL. This mainly benefits from our solution to both label-inconsistency and label-domination problem. Besides, we show that the images generated from Relaxed cGAN are of correct semantics and higher quality, compared to those from the baselines in Fig.~\ref{fig:label-inconsistency} and Appendix D. Such results agree with the classification performance and well support our motivation.
\textbf{DomainNet \& Office-Home.}
We further evaluate Relaxed cGAN on two large-scale image datasets, DomainNet and Office-Home. We conduct all experiments under the 3-shot setting with a pretrained Alexnet backbone. We choose the label data in the target domain following the setting of ~\cite{saito2019semi}.

As shown in Table~\ref{tab:domainnet-experiment}, on DomainNet our method surpasses ENT~\cite{grandvalet2005semi} by a large margin, which uses the same loss function as the classifier in Relaxed cGAN. This suggests that thanks to the carefully designed generator and discriminator, the transferred images can provide useful learning signals to the classifier. Further, as a pixel-level method, Relaxed cGAN achieves SOTA results on DomainNet demonstrating its scalability on large datasets. Similar results are obtained on Office-Home referring to Table \ref{tab:office-home}, where Relaxed cGAN outperforms existing powerful methods.
% Last but not the least, we visualize the samples from Relaxed cGAN on DomainNet in Fig.~\ref{fig5:DomainNet}. The transferred images share the same semantics with the input but has different domain styles, demonstraing the effectiveness of Relaxed cGAN. 

% Table generated by Excel2LaTeX from sheet 'total'
% \begin{table}[htbp]
%   \centering
%     \begin{tabular}{lrrrrrrr}
%     \toprule
%           & \multicolumn{1}{l}{R2C} & \multicolumn{1}{l}{R2P} & \multicolumn{1}{l}{P2C} & \multicolumn{1}{l}{C2S} & \multicolumn{1}{l}{S2P} & \multicolumn{1}{l}{R2S} & \multicolumn{1}{l}{P2R} \\
%     \midrule
%     S+T   & 47.1  & 45    & 44.9  & 36.4  & 38.4  & 33.3  & 58.7 \\
%     ENT   & 45.5  & 42.6  & 40.4  & 31.1  & 29.6  & 29.6  & 60 \\
%     MME   & 55.6  & 49    & 51.7  & 39.4  & 43    & 37.9  & 60.7 \\
%     Relaxed cGAN &       &       &       & \textbf{44.1} &       & \textbf{41.3} &  \\
%     \bottomrule
%     \end{tabular}%
%   \caption{DomainNet results. All experiment are conducted under 3-shot setting with Alexnet backbone.}
%   \label{tab:addlabel}
% \end{table}

\begin{table}[t]
  \centering
    \footnotesize{\setlength{\tabcolsep}{1.8mm}{\begin{tabular}{lccccccc}
    \toprule
          & \multicolumn{1}{l}{Relaxed} & \multicolumn{1}{l}{BiAT} & \multicolumn{1}{l}{APE} & \multicolumn{1}{l}{meta-} & \multicolumn{1}{l}{MME} & \multicolumn{1}{l}{ENT} & \multicolumn{1}{l}{S+T} \\
          & \multicolumn{1}{l}{cGAN} & ~\cite{jiangbidirectional} & ~\cite{kim2020attract} & MME~\cite{li2020online} & ~\cite{saito2019semi} & ~\cite{grandvalet2005semi} &  \\
    \midrule
    R to C   &  56.8          & \textbf{58.6} & 54.6 & 56.4    & 55.6  & 45.5  & 47.1 \\
    R to P   &  \textbf{51.8} & 50.6  & 50.5 & 50.2   & 49.0    & 42.6  & 45.0 \\
    P to C   &  52.0          & 52.0  & \textbf{52.1} & 51.9   & 51.7  & 40.4  & 44.9 \\
    C to S   &  \textbf{44.1} & 41.9  & 42.6 & 39.6 & 39.4 & 31.1  & 36.4 \\
    S to P   &  \textbf{44.2} & 42.1  & 42.2 & 43.7 & 43.0 & 29.6  & 38.4 \\
    R to S   &  \textbf{42.8} & 42.0  & 38.7 & 38.7 & 37.9 & 29.6  & 33.3 \\
    P to R   &  \textbf{61.1} & 58.8  & 61.4 & 60.7 & 60.7 & 60.0  & 58.7 \\
    AVG      &  \textbf{50.5} & 49.4  & 48.9 & 48.8 & 48.2 & 39.8  & 43.4  \\
    \bottomrule
    \end{tabular}}}
\caption{DomainNet results. All experiments are conducted under the 3-shot setting with an Alexnet backbone. Our method outperforms baselines in most settings and achieves SOTA on the average results. }
\label{tab:domainnet-experiment}
\vspace{-0.2cm}
\end{table}%

\begin{table}[t]
\centering
\footnotesize{\setlength{\tabcolsep}{1.8mm}{\begin{tabular}{ccccccc}
\toprule
 & Relaxed & APE & meta- & MME & ENT & S+T \\
 & cGAN & ~\cite{kim2020attract} & MME~\cite{li2020online} & ~\cite{saito2019semi} & ~\cite{grandvalet2005semi} & \textbf{} \\
 \midrule
R to C & \textbf{68.4} & 66.4 & 65.2 & 64.6 & 62.6 & 55.7 \\
R to P & 85.5 & \textbf{86.2} & - & 85.5 & 85.7 & 80.8 \\
R to A & \textbf{73.8} & 73.4 & - & 71.3 & 70.2 & 67.8 \\
P to R & 81.2 & \textbf{82.0} & - & 80.1 & 79.9 & 73.1 \\
P to C & \textbf{68.1} & 65.2 & 64.5 & 64.6 & 60.5 & 53.8 \\
P to A & \textbf{67.9} & 66.1 & 66.7 & 65.5 & 63.9 & 63.5 \\
A to P & 79.0 & \textbf{81.1} & - & 79.0 & 79.5 & 73.1 \\
A to C & \textbf{64.3} & 63.9 & 63.3 & 63.6 & 61.3 & 54.0 \\
A to R & 80.1 & \textbf{80.2} & - & 79.7 & 79.1 & 74.2 \\
C to R & \textbf{77.5} & 76.8 & - & 76.6 & 76.4 & 68.3 \\
C to A & 66.3 & 66.6 & \textbf{67.5} & 67.2 & 64.7 & 57.6 \\
C to P & 78.3 & \textbf{79.9} & - & 79.3 & 79.1 & 72.3 \\
AVG & \textbf{74.2} & 74.0 & - & 73.1 & 71.9 & 66.2 \\
\bottomrule
\end{tabular}}}
\caption{Office-Home results under 3-shot setting with an ResNet-34 backbone. Relaxed cGAN outperforms baselines in most settings and achieves SOTA on the average results.}
\label{tab:office-home}%
\end{table}

\subsection{Ablation Study}
\label{sec:ablation-study}

% Table generated by Excel2LaTeX from sheet '表格'
\begin{table}[t]
  \centering
    \footnotesize{\begin{tabular}{lccc}
    \toprule
          & \multicolumn{1}{l}{$\mathcal{S}\rightarrow \mathcal{M}$} & \multicolumn{1}{l}{$\mathcal{M}\rightarrow \mathcal{U}$} & \multicolumn{1}{l}{$\mathcal{U}\rightarrow \mathcal{M}$} \\
    \midrule
    ENT          & 83.9  &  85.1      &  83.9 \\
    cycleGAN                & 68.4  &  95.9      &  94.2 \\
    Preliminary Method  & 87.8  &  82.1     &  84.9 \\
    Relaxed cGAN w/o. & \multirow{2}[2]{*}{94.3}  &  \multirow{2}[2]{*}{96.4}     &  \multirow{2}[2]{*}{96.2} \\
    \quad$\mathcal{L}_{marg}$, $\mathcal{L}_{pseudo}$ & & & \\
    \midrule
    Relaxed cGAN            & \textbf{95.9}  &  \textbf{97.9}      &  \textbf{98.7} \\
    \bottomrule
    \end{tabular}}
\caption{Ablation Study. We conduct all the experiments under the 1-shot setting on 3 adaptation tasks. \textbf{ENT}: semi-supervised learning on both labeled and unlabeled target data with $\mathcal{L}_{ENT}$. \textbf{Relaxed cGAN w/o. $\mathcal{L}_{marg}$, $\mathcal{L}_{pseudo}$}: Relaxed cGAN without the $\mathcal{L}_{marg}$ amd $L_{pseudo}$.}
\label{tab:ablation}
\vspace{-0.2cm}
\end{table}%

% \begin{table*}[htbp]
%   \centering
%     \begin{tabular}{lrr|rr|rr}
%     \toprule
%           & \multicolumn{2}{c}{s2m} & \multicolumn{2}{c}{u2m} & \multicolumn{2}{c}{m2u} \\
%           & \multicolumn{1}{l}{1-shot} & \multicolumn{1}{l}{3-shot} & \multicolumn{1}{l}{1-shot} & \multicolumn{1}{l}{3-shot} & \multicolumn{1}{l}{1-shot} & \multicolumn{1}{l}{3-shot} \\
%     \midrule
%     SSL   &       &       &       &       &       &  \\
%     cycleGAN &       &       &       &       &       &  \\
%     Relaxed cGAN w.o tricks &       &       &       &       &       &  \\
%     Relaxed cGAN w. y in G &       &       &       &       &       &  \\
%     Relaxed cGAN &       &       &       &       &       &  \\
%     \bottomrule
%     \end{tabular}%
%   \caption{Add caption}
%   \label{tab:addlabel}%
% \end{table*}%

% add accuracy comparison (different architecture; spectral norm; hyper-parameters)
We conduct several quantitative ablation studies to analyze the contribution of each component on 
% $\mathcal{S} \rightarrow \mathcal{M}$, $\mathcal{M} \rightarrow \mathcal{U}$, $\mathcal{U} \rightarrow \mathcal{M}$ 
digit benchmarks in the high-resource SSDA settings, with 1 annotated samples for each class. The results are shown in Table~\ref{tab:ablation}. 

\textbf{Label-inconsistency and Label-domination.} As shown in Table~\ref{tab:ablation}, cycleGAN encounters label-inconsistency problem when the domain shift is large ($\mathcal{S} \rightarrow \mathcal{M}$). The Preliminary Method encounters the label-domination problem and only shows comparable results as the entropy-minimization (ENT) on the target domain. The proposed Relaxed cGAN resolves both problems and surpasses the above two approaches by a large margin.

% \textbf{Label-inconsistency.} When the domain shift is large ($\mathcal{S} \rightarrow \mathcal{M}$), cycleGAN encounters label-inconsistency problem. It generates label-inconsistent samples and introduces wrong training signals, resulting in poor classification results. 

% \textbf{Label-domination.} The Preliminary Method tends to neglect the source images' information, which prevents from an effective domain adaptation. As we can see, it shows only marginal improvement over the entropy-minimization (ENT) on target domain. 

% When domain shift is relatively small ($\mathcal{M} \rightarrow \mathcal{U}$, $\mathcal{U} \rightarrow \mathcal{M}$), preliminary method can not transfer diverse images due to label-domination, resulting in similar results as semi-supervised learning on target domain (ENT). Relaxed cGAN outperform both, validating that label-inconsistency and label-domination are both effectively addressed.
\textbf{Additional Regularization.}
We show the effectiveness of both marginal loss and pseudo loss together as they all try to make better use of unlabeled data. As we can see from the last two rows of Table~\ref{tab:ablation}, these additional regularizations further boost Relaxed cGAN's performance on SSDA.

\section{Conclusion}

In this work, we focus on the setting of SSDA. We identify the label-domination problem raised by the conditional GAN framework during the image transformation.
% , including the label-inconsistency problem and the label-domination problem.
% These two problems usually result in poor source image transfer and finally hurt the performance of the classification.
We then elaborately design Relaxed cGAN to resolve this issue.
Its generator takes images as the only input, while its discriminator takes image-label pairs. By doing this, the generator has to infer the input data's semantic information. 
The proposed model has a provable and satisfied theoretical equilibrium. Its practical convergence property and image transfer quality are also empirically shown to be effective. 
% We  elaborately design Relaxed cGAN, which consists of a conditional discriminator, an unconditional generator, as well as a jointly trained classifier, to alleviate the two problems and eventually achieve satisfactory classification accuracy.
% Empirically, Relaxed cGAN is able to generate diverse data in the target domain through adaptation, which leads to substantial improvements in classification.
Extensive experimental results show that the proposed Relaxed cGAN achieves state-of-the-art results in DomainNet and several digit adaptation benchmarks under both low-resource and high-resource SSDA settings. The model also obtains competitive results in the Office-Home benchmark.

\section*{Aknowledgement}

This work was supported by the National Key Research and Development Program of China (Nos. 2017YFA0700904, 2020AAA0104304), NSFC Projects (Nos. 61620106010, 62076145, U19B2034, U1811461, U19A2081), Beijing NSF Project (No. L172037), Beijing Academy of Artificial Intelligence (BAAI), THU-Bosch JCML center, Tsinghua-Huawei Joint Research Program, a grant from Tsinghua Institute for Guo Qiang, Tiangong Institute for Intelligent Computing, the JP Morgan Faculty Research Program and the NVIDIA NVAIL Program with GPU/DGX Acceleration. C. Li was supported by the Chinese postdoctoral innovative talent support program and Shuimu Tsinghua Scholar.

%%%%%%%%% BODY TEXT
{\small
\bibliographystyle{ieee_fullname}
\bibliography{rcgan}
}

\clearpage

% \input{appendix}
% \appendix



\section*{Supplementary material}

# Supplementary Material for "Relaxed Conditional Image Transfer for Semi-supervised Domain Adaptation"

## 1 Appendix A: Equilibrium Analysis

For the ease of the proof, we define our objective function for the game of three networks $C$, $G_{S \to T}$ and $D_T$ as:

$$\min_{C, G_{S \to T}} \max_{D_T} \tilde{U}(C, G_{S \to T}, D_T) = \min_{C, G_{S \to T}} \max_{D_T} \mathcal{L}_{GAN}(C, G_{S \to T}, D_T) + \min_C \mathcal{L}_C, \quad (1)$$

where

$$\begin{aligned}
\mathcal{L}_{GAN}(C, G_{S \to T}, D_T) &= \mathbb{E}_{(x,y) \sim p_t(x,y)}[\log(D_T(x,y))] \\
&+ \alpha \mathbb{E}_{(x,y) \sim p_s(x,y)}[\log(1 - D_T(G_{S \to T}(x), y))] \\
&+ ((1-\alpha)\mathbb{E}_{(x,y) \sim p_s(x,y)}[\log(1 - D_T(x, C(x)))], \quad (2)
\end{aligned}$$

refer to $\mathcal{L}_{GAN}$ and $\mathcal{L}_C$ in Sec.4.

We first give the proof that the equilibrium of $\tilde{U}(C, G_{S \to T}, D_T)$ is achieved only when $p_t(x,y) = p_g(x,y) = p_c(x,y)$, under the nonparametric assumption [1]. After that, a corollary is shown that adding other losses, such as $L_{cycle}$, $L_{marg}$, $L_{ent}$, will not change such equilibrium of our framework.

**Lemma 1.1.** *For any fixed $C$ and $G_{S \to T}$, the optimal discriminator $D_T$ of the game defined by the objective function $U(C, G_{S \to T}, D_T)$ is*

$$D^*_{T|C, G_{S \to T}}(x, y) = \frac{p_t(x,y)}{p_t(x,y) + p_m(x,y)}, \quad (3)$$

*where $p_m(x,y) := \alpha p_g(x,y) + (1-\alpha) p_c(x,y)$ is a mixture distribution of $p_g(x,y)$ and $p_c(x,y)$.*

*Proof.* Given the classifier and generator, the objective function can be rewritten as

$$\begin{aligned}
U(C, G_{S \to T}, D_T) &= \iint p_t(x,y) \log D_T(x,y) dy dx \\
&+ \alpha \iint p(y) p_s(x) \log(1 - D_T(G_{S \to T}(x), y)) dy dx \\
&+ (1-\alpha) \iint p_t(x) p_c(y|x) \log(1 - D_T(x,y)) dy dx \\
&= \iint p_t(x,y) \log D_T(x,y) dy dx \\
&+ \iint p_m(x,y) \log(1 - D_T(x,y)) dy dx,
\end{aligned}$$

which achieves the maximum at $\frac{p_t(x,y)}{p_t(x,y) + p_m(x,y)}$.

□

Given $D^*_{T|C,G_{S \to T}}$, we can reformulate the minimax game with value function $U$ as:

$$V(C, G_{S \to T}) = \iint p_t(x,y) \log \frac{p_t(x,y)}{p_t(x,y) + p_m(x,y)} dy dx$$
$$+ \iint p_m(x,y) \log \frac{p_m(x,y)}{p_t(x,y) + p_m(x,y)} dy dx.$$

**Lemma 1.2.** *The global minimum of $V(C, G_{S \to T})$ is achieved if and only if $p_t(x,y) = p_m(x,y)$.*

*Proof.* Following the proof in GAN [1], the $V(C, G_{S \to T})$ can be rewritten as

$$V(C, G_{S \to T}) = -\log 4 + 2\mathbb{D}_{JS}(p_t(x,y) || p_m(x,y)), \tag{4}$$

where $\mathbb{D}_{JS}$ is the Jensen-Shannon divergence, which is always non-negative and the unique optimum is achieved if and only if $p_t(x,y) = p_m(x,y)$. □

**Theorem 1.3.** *The equilibrium of $\tilde{U}(C, G_{S \to T}, D_T)$ is achieved if and only if $p_t(x,y) = p_g(x,y) = p_c(x,y)$.*

*Proof.* According to the definition, we have $\tilde{U}(C, G_{S \to T}, D_T) = \mathcal{L}_{GAN}(C, G_{S \to T}, D_T) + \mathcal{L}_C$. We can rewrite $\mathcal{L}_C$ as:

$$\mathcal{L}_C = \mathbb{E}_{p_t(x,y)}[-\log p_c(y|x)]$$
$$= \mathbb{E}_{p_t(x,y)}[\log \frac{p_t(x,y)}{p_c(x,y)}] - \mathbb{E}_{p_t(x,y)}[\log p(y|x)]$$
$$= \mathcal{D}(p_t(x,y) || p_c(x,y)) + \mathbb{E}_{p_t(x)}[\mathcal{H}[p(y|x)]],$$

where $\mathcal{D}(\cdot || \cdot)$ denotes the KL-divergence and $\mathcal{H}[\cdot]$ denotes the differential entropy. Since the second term is a constant and determined by the data distribution, minimizing $\mathcal{L}_C$ is equivalent to minimizing $\mathcal{D}(p_t(x,y) || p_c(x,y))$, which is always non-negative and zero if and only if $p_t(x,y) = p_c(x,y)$. Besides, the previous lemmas can also be applied to $\tilde{U}(C, G_{S \to T}, D_T)$, which indicates that $p_t(x,y) = p_m(x,y)$ at the global equilibrium and concludes the proof. □

**Corollary 1.3.1.** *Adding any divergence (e.g. the KL divergence) between any two of the joint distributions or the conditional distributions or the marginal distributions to $\tilde{U}$ as an additional regularization to be minimized, will not change the global equilibrium of $\tilde{U}$.*

*Proof.* This conclusion can be straightforwardly obtained by the global equilibrium point of $\tilde{U}$ and the definition of a statistical divergence between two distributions. □

## 2 Appendix B: Network Architecture

Since the digit dataset SVHN($\mathcal{S}$) is more complex than MNIST($\mathcal{M}$) and USPS($\mathcal{U}$). We employ different architectures when transferring between the datasets including $\mathcal{S}$ or not. For adaptation between $\mathcal{U}$ and $\mathcal{M}$, a simple architecture is utilized, refer to the Simple G and Simple D in Fig. 1. The Complex G and Complex D of Fig. 1 are adopted in the transfer tasks from $\mathcal{S}$ to $\mathcal{M}$ and $\mathcal{S}$ to $\mathcal{U}$. For DomainNet [6], we set the default setting of cycleGAN [8] as our base architecture, with modification of Spectral Normalization and the suitable hinge loss [5]. Alexnet [4] is our classifier following the setting of MME [7]. We implement the conditional discriminator by outputting a (K+1)-dim vector of which each entry is the logit of the corresponding class. K is the number of classes, and the class K+1 refers to the marginal loss, as discussed in Sec.3.1.

## 3 Appendix C: Label Domination Study

We conduct a series of experiments to demonstrate the existence of the label-domination problem. The first two sets of experiments show how we identify such a problem. The latter three experiments indicate that such a problem still exists even when we employ cycle consistency loss [8], complex architecture, and Spectral Normalization [5] to the original model, which are normally considered

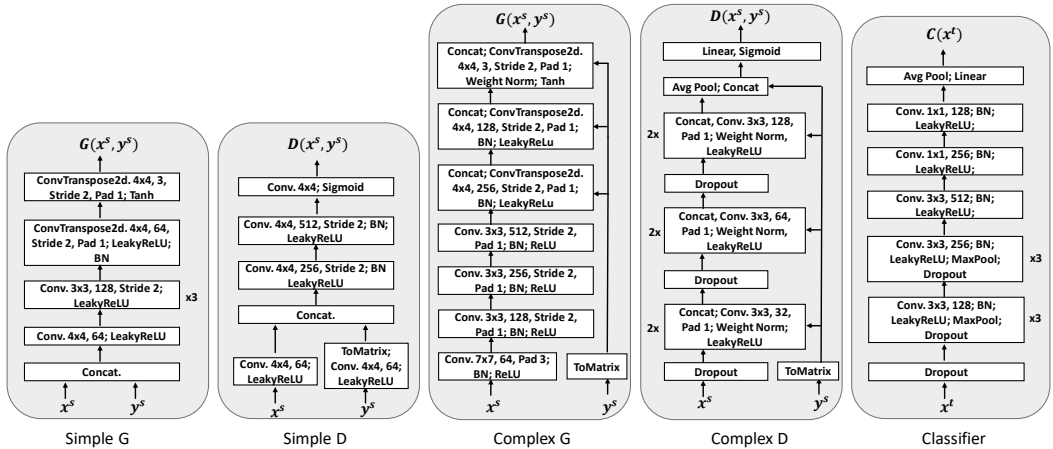

Figure 1: Network Architectures for digit datasets. "ToMatrix" means that each label is expanded to a $K*h*w$ feature map. The $i$-th channel is all set to 1, others are 0. $K$ is the number of classes, $i$ is the value of the label, $h$ and $w$ are the height and width of the image respectively.

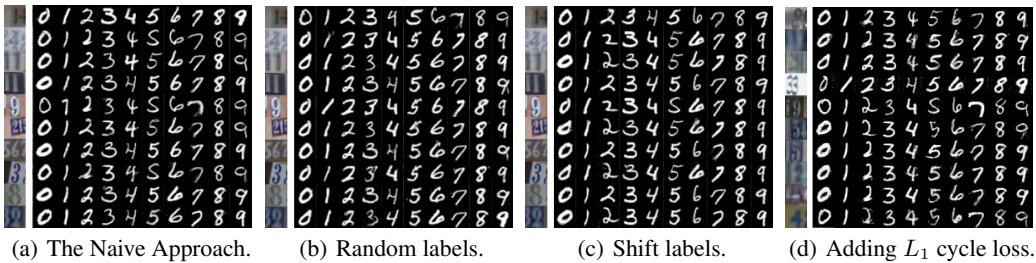

(a) The Naive Approach.  (b) Random labels.  (c) Shift labels.  (d) Adding $L_1$ cycle loss.

Figure 2: Label-domination problem. In each panel, the generator receives images from the leftmost column, paired with each label from '0' to '9'. In the right part, the image in the $i$-th row and $j$-th column is the corresponding generated image of the $i$-th input image and label '$j$'. (a) The Naive Approach: the conditional generator and the conditional discriminator with correct image-label pairs are input to $G$ during training stage. (b) Source images with random labels are input to $G$. (c) Source images with shift labels ('1' to '2', '2' to '3', and so on) as input to $G$. (d) The Naive Approach with cycle-consistency loss. The label-domination problem occurs in all settings.

as useful ways to solve the problem. All experiments are conducted in $\mathcal{S} \rightarrow \mathcal{M}$, following the description of Sec.3.2.

**Inference with incorrect labels.** We use the straight-forward approach with the conditional generator and the conditional discriminator described in Sec.3.2 for training. However, during inference, we feed source images with different labels to see if the information of the source images is taken into account during the transferring procedure. The results in Fig. 2 (a) show that the generated images completely follow the input label and ignore the source images' information.

**Training with random and shifted labels.** We further construct two validation approaches to better show that the source image is indeed be ignored during transfer. The generator is fed a source image with a random label or a label shifted by one (e.g., from '1' to '2' and '2' to '3' and so on) respectively in model training(namely, we input wrong image-label pairs to the generator). The discriminator still accepts the real label-data pairs in the target domain. Again, during inference, we feed source images with different labels, and the results are shown in Fig. 2 (b-c). We observe the same phenomenon of these two settings as the naive approach that the label-domination still holds. Besides, by adopting our jointly-training framework for SSDA, the accuracy of the Naive Approach, Random labels, and Shift labels are close to each other, with 88.0%, 87.9%, and 87.5% respectively. The generation and the accuracy indirectly show that the semantic information of the source image itself is ignored.

3

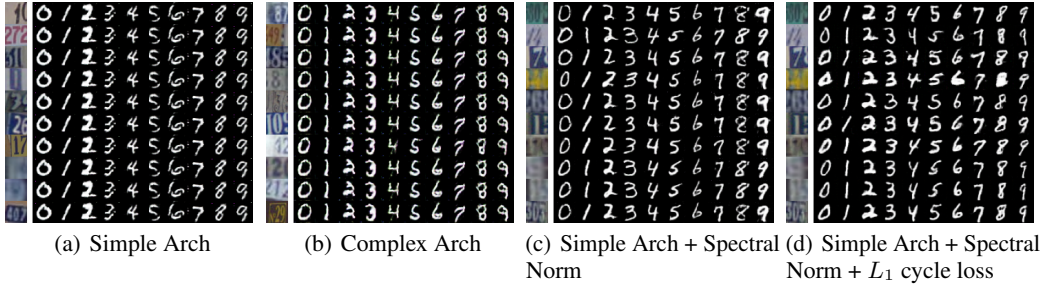

| (a) Simple Arch | (b) Complex Arch | (c) Simple Arch + Spectral Norm | (d) Simple Arch + Spectral Norm + $L_1$ cycle loss |

Figure 3: Label Domination Problem still occurs with a complex architecture, Spectral Normalization and cycle consistency loss.

**Cycle Consistency Loss.** Cycle-consistency loss is a widely adopted loss for maintaining the pixel-level correlation between the source image and the transferred image, which is proposed by [8]. We apply the cycle-consistency loss to the straight-forward approach to see if the label domination problem can be relieved. The results are presented in Fig. 3(d). As we can see, the label domination problem still exists.

**Different Model Architectures.** To eliminate the impact of model architecture, we explore two different architectures of generator and discriminator, namely, simple and complex, as specified in Fig. 1. We follow the same experimental protocol in the paper, i.e. we feed source images with different labels to see if the source images' semantic information is taken into account during the transferring procedure. The results are shown in Fig. 3(a) and Fig. 3(b). As we can see, the label domination problem occurs in both architectures. However, we find that mode collapse also appears, i.e. identical images are generated for each input label, regardless of source images. Therefore, we apply Spectral Normalization [5] to discriminator to stabilize training.

**Spectral Normalization.** Spectral Normalization is a weight normalization approach to stabilize the training of discriminator. It mitigates the exploding gradient and mode collapse problem by controlling the Lipschitz constant of the discriminator. Readers may refer to [5] for more detail. Trying to eliminate the effect of label domination, we substitute all normalization layers (such as batch normalization) in the discriminator with Spectral Normalization. The result is shown in Fig. 3(c). Evidently, the mode collapse problem is resolved. However, we can see that label still dominates the generation, which is not satisfactory. If not specified, all models in the paper adopt cycle consistency loss and Spectral Normalization.

## 4 Appendix D: Additional Results

### 4.1 Unlabeled data in target domain

| Dataset | # of data per class | ACAL | Relaxed cGAN |
|---|---|---|---|
| $\mathcal{S} \to \mathcal{M}$ | 10 | 93.9 | **94.9** |
| $\mathcal{S} \to \mathcal{M}$ | 100 | 91.8 | **97.3** |
| $\mathcal{S} \to \mathcal{M}$ | 6000 | 92.2* | **99.1** |

Table 1: Experiments of unlabeled data in the target domain. The number of labeled samples per class is fixed as 10. The result marked by* is obtained in our implementation and others are from the ACAL paper [3].

We show that our jointly-training framework also makes good use of unlabeled data. We fix the number of labeled samples to be 10 for each class and increase the number of unlabeled ones. The results are summarized in Table 1. Our model's performance continuously increases as we add more unlabeled data, indicating its effective utilization of these unlabeled samples. We mainly contribute this to the adoption of the marginal loss and the ENT loss [2].

## 4.2 Quality of generated images

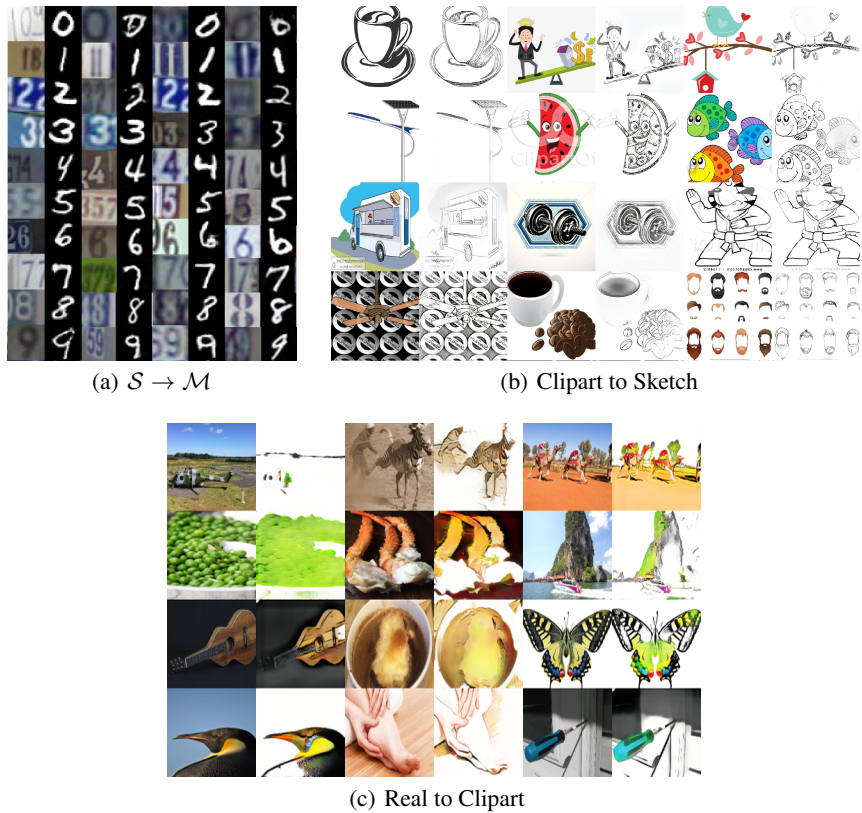

(a) $\mathcal{S} \rightarrow \mathcal{M}$

(b) Clipart to Sketch

(c) Real to Clipart

Figure 4: Visualization of $\mathcal{S} \rightarrow \mathcal{M}$ and DomainNet generation. The odd columns are the inputs to the generator, and the even columns are the corresponding outputs.

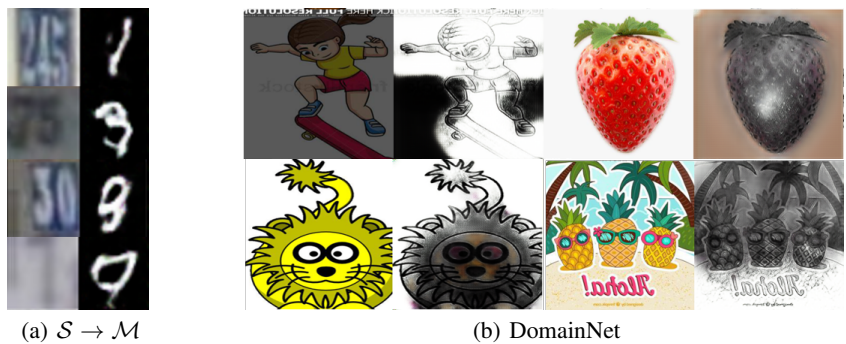

(a) $\mathcal{S} \rightarrow \mathcal{M}$

(b) DomainNet

Figure 5: Visualization of some failure cases.

In the paper, we show several representative figures to demonstrate our idea. Here we put more generation results to show the diversity and output quality of Relaxed cGAN, referring to Fig. 4. We also illustrate several badly generated samples of Relaxed cGAN, as shown in Fig. 5. We take the assumption that the existence of these failure cases is mainly caused by the unstable training of GAN, which may a future direction of considering GAN training into SSDA.

5



\end{document}